%% file: main.tex
\documentclass[10pt,twocolumn,letterpaper]{article}
\usepackage[T1]{fontenc}
\usepackage[pagenumbers]{cvpr}
\input{preamble}
\definecolor{paperblue}{rgb}{0.21,0.49,0.74}
\usepackage[pagebackref,breaklinks,colorlinks,allcolors=paperblue]{hyperref}
\hypersetup{
    pdftitle={FujinSplat: Seeing Through Smoke with RAW-Domain Novel-View Synthesis},
    pdfauthor={Gengjia Chang, Ziteng Cui, Shuhong Liu},
    pdfkeywords={}
}
\title{FujinSplat: Seeing Through Smoke with RAW-Domain Novel-View Synthesis}
\newcommand{\correspondencemark}{\ensuremath{\lozenge}}
\author{
    Gengjia Chang\textsuperscript{1} \qquad Ziteng Cui\textsuperscript{2} \qquad Shuhong Liu\textsuperscript{2,\correspondencemark}\\
    \textsuperscript{1}Hefei University of Technology \qquad \textsuperscript{2}The University of Tokyo
}
\begin{document}
\maketitle
\begingroup
    \renewcommand{\thefootnote}{\correspondencemark}
    \footnotetext[1]{Corresponding author.}
\endgroup
\begin{strip}
    \captionsetup{hypcap=false}
    \centering
    \includegraphics[width=\textwidth]{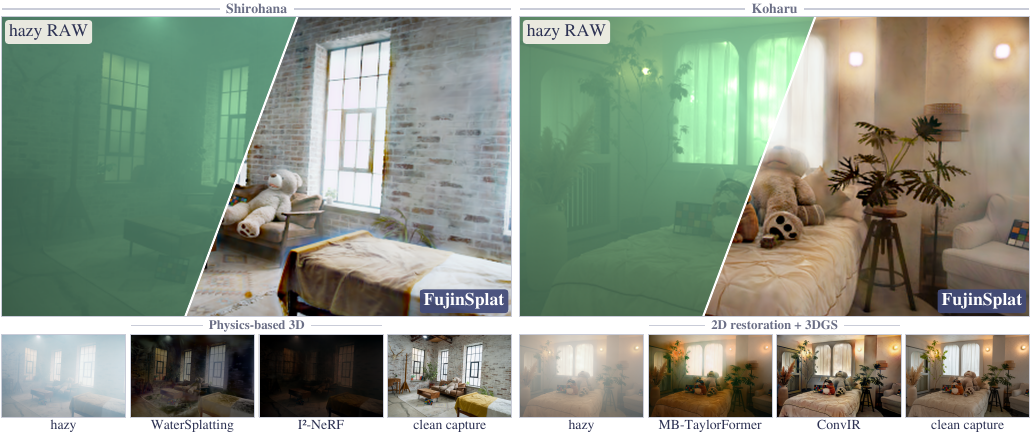}
    
    \captionof{figure}{\textbf{FujinSplat} on RealX3D smoke scenes~\cite{realx3d}.
    \textbf{\textit{Top:}} our novel view right of the diagonal and the hazy RAW of the nearest training view left of it, as novel views are recorded only in clean air.
    \textbf{\textit{Bottom:}} the same view from physics-based 3D methods~\cite{watersplatting,liu2025i2nerf} and restoration networks~\cite{mbtaylorformer,convir} with 3DGS.}
    
    \label{fig:teaser}
\end{strip}
\input{sec/0_abstract}
\input{sec/1_intro}

\input{sec/2_related_work}
\input{sec/3_method}
\input{sec/4_experiments}

\input{sec/5_conclusion}

{
    \small
    \bibliographystyle{ieeenat_fullname}
    \bibliography{main}
}
\input{sec/X_suppl}
\end{document}

%% file: preamble.tex
\usepackage{fix-cm}
\usepackage{microtype}
\usepackage{colortbl}
\usepackage{cuted}
\usepackage{multirow}
\definecolor{fst}{HTML}{9FC5E8}
\definecolor{sed}{HTML}{C9DCF0}
\definecolor{thd}{HTML}{E1E0F2}
\definecolor{yesblue}{HTML}{4F86B5}
\definecolor{nored}{HTML}{C66A6A}

\newcommand{\nomark}{\textcolor{nored}{\ensuremath{\times}}}
\newcommand{\gptmodel}[1]{%
    \mbox{\raisebox{-0.12em}[0pt][0pt]{\includegraphics[height=0.90em]{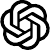}}%
        \kern0.12em-\kern0.12em #1}}
\newcommand{\nanobananamodel}[1]{%
    \mbox{\raisebox{-0.12em}[0pt][0pt]{%
        \includegraphics[height=0.90em]{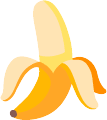}}%
        \kern0.12em-\kern0.12em #1}}
\makeatletter
\renewcommand\paragraph{\@startsection{paragraph}{4}{\z@}{6pt plus 2pt minus 2pt}{-1em}{\normalfont\normalsize\bfseries}}
\makeatother

%% file: sec/0_abstract.tex


\begin{abstract}
Novel-view synthesis through smoke must reconstruct a scene from input views in which the smoke attenuates the scene radiance and adds light scattered toward the camera. Existing dehazing and medium-aware reconstruction methods model this scattered light as a uniform airlight and attribute the spatial variation of the haze to scene depth. We analyze real indoor smoke in linear RAW domain, where pixel values are proportional to the incident light, and find that this assumption does not hold. The transmission is nearly uniform within each view, and the spatial variation of the smoke instead comes from the scattered light, which follows local light sources in both distribution and color. Its share of the recorded light is set by how brightly the smoke is lit relative to the scene. In linear RAW, this structure allows the smoke to be removed by rescaling the recorded signal rather than by synthesizing content. We therefore propose \textbf{FujinSplat}, a RAW-domain Gaussian splatting framework for novel-view synthesis through indoor smoke. For each view, a network predicts a low-resolution gain field that removes the scattered light and a single transmission that restores the attenuated detail. The corrected views remain physically valid linear images, and Gaussian splatting fuses them while averaging out the sensor noise that the detail restoration amplifies. The network is trained exclusively on a simulated dataset that reproduces these lighting effects. On real-world RealX3D dataset, FujinSplat achieves a novel-view PSNR of $18.88$\,dB and outperforms the strongest non-generative baseline by $2.80$\,dB. Code is available at \url{https://i2wm.github.io/FujinSplat/}.
\end{abstract}

%% file: sec/1_intro.tex
\section{Introduction}
\label{sec:intro}

Novel-view synthesis through smoke aims to render the clean scene from input views degraded by the smoke. Since radiance fields and 3D Gaussian splatting~\cite{mildenhall2020nerf,kerbl3Dgaussians,li2026densesplat} reproduce the appearance of their input views~\cite{deraings,cui2024alethnerf,luminancegspp}, existing methods either restore each view with a dehazing network before reconstruction~\cite{psd,promptir,moceir,mbtaylorformer,convir} or model the scattering medium jointly with the scene~\cite{chen2023dehazenerf,ramazzina2023scatternerf,seathrunerf,watersplatting,seasplat,dehazegs2025,liu2025i2nerf}. Most of them rely, explicitly or through their training data, on the atmospheric scattering model~\cite{narasimhan2002vision,he2011single},
\begin{equation}
    I(\mathbf{x}) = J(\mathbf{x})\,t(\mathbf{x}) + A\,\bigl(1-t(\mathbf{x})\bigr),
    \quad t(\mathbf{x}) = e^{-\beta d(\mathbf{x})},
    \label{eq:asm}
\end{equation}
where $J$ is the clean radiance, $A$ is a single airlight, and the transmission $t$ decays with scene depth $d$ at density $\beta$. This model assumes that a single airlight illuminates the medium uniformly and attributes the spatial variation of the haze to scene depth. The assumption is appropriate for daytime haze under sky illumination~\cite{koschmieder1924theorie,lee2016visibility,zhang2017fast,yan2020nighttime}, and most synthetic training data for dehazing are generated with Eq.~\eqref{eq:asm} or with extensions that retain its single airlight~\cite{li2019benchmarking,wu2023ridcp}. Methods for smoke inherit this haze model, since they reuse dehazing networks, dehazing training data, or the same image formation.

The assumption of a uniform airlight fails for indoor smoke, which is illuminated by the local light sources of the room rather than by the sky. Its scattered light therefore peaks around these sources rather than where the scene is deepest. Depth-based methods thus leave a residual veil around the light sources. In Fig.~\ref{fig:teaser}, WaterSplatting~\cite{watersplatting} darkens the entire room of \textsc{Shirohana} while its windows remain veiled. How much light the smoke adds also depends on the room lighting, whereas synthetic dehazing data draw the airlight independently of the scene~\cite{li2019benchmarking,wu2023ridcp}.

The processing of the camera further obscures both the shape and the level of the light scattered by the smoke. Eq.~\eqref{eq:asm} describes scene radiance, whereas existing methods typically operate on images rendered by the image signal processor (ISP). Its tone mapping and gamma encoding are nonlinear~\cite{liu2026rawvla} and break the additive relation between the attenuated scene radiance and the scattered light. The ISP also clips the light sources and suppresses the low-contrast detail that remains behind the smoke. A linear RAW image preserves both, since its pixel values are proportional to the incident light up to the white level~\cite{cui2024rawadapter,huang2025drraw}. Its sensor noise is also zero-mean with a known variance~\cite{liu2026deepsky}.

We quantify how indoor smoke behaves in linear RAW on $168$ pairs of captures taken with and without smoke from RealX3D~\cite{realx3d}. A model with one transmission per view and a spatially varying veil fits every view better than Eq.~\eqref{eq:asm} with a free transmission map and improves the median fit by $6.1$\,dB. The residual between the smoke and clean captures concentrates $98.7\%$ of its energy along a single color direction. The light on the smoke accounts for $78\%$ of the spatial variation of the veil, and the veil takes the color of the light sources. The fraction of the observed light that must be removed is set by how brightly the smoke is lit relative to the scene, a quantity we call the light ratio. Across the eight rooms, this fraction correlates with the light ratio at a Spearman coefficient of $0.90$ and with the optical thickness of the smoke at only $0.19$ (Sec.~\ref{sec:formation}).

Because these relations are linear in RAW, the smoke can be removed by rescaling what the camera recorded rather than by synthesizing content. We build \textbf{FujinSplat}, a RAW-domain Gaussian splatting framework for novel-view synthesis through smoke, on this principle. For each RAW view, a network predicts a low-resolution RGB gain field, which removes the smooth veil, and a single transmission, which restores the attenuated detail. Fixed operators apply both quantities to the observation, so each corrected view remains a linear RAW image that contains only what the camera recorded. 3D Gaussian splatting~\cite{kerbl3Dgaussians} then reconstructs the scene from these views in linear RAW. Because the sensor noise in RAW has a known variance, the operator amplifies detail only where it exceeds this noise, and the reconstruction averages the remaining noise across views. Both predicted quantities are physical and can therefore be supervised with simulated labels. This supervision is essential. Trained through a pixel loss alone, the transmission settles near $0.5$ instead of its true value of $0.23$, and the detail gain stays below half of its physical value. We train the network solely on a light-aware simulator whose veil follows the light sources in shape and the light ratio in level.

On the eight smoke scenes of RealX3D, we compare FujinSplat with physics-based 3D methods, with restoration networks followed by the same Gaussian splatting pipeline, and with hybrid pipelines that use generative image models. FujinSplat reaches an average novel-view PSNR of $18.88$\,dB, which is $2.80$\,dB above ConvIR~\cite{convir}, the strongest baseline in Table~\ref{tab:main_results}. It also achieves the best average SSIM and LPIPS. Our contributions are summarized as follows.
\begin{itemize}
    \item A measurement on real paired RAW captures showing that indoor smoke is governed by its illumination.
    \item A light-aware indoor smoke simulator that reproduces these measurements in camera RAW.
    \item FujinSplat, a RAW-domain Gaussian splatting framework for novel-view synthesis through smoke, whose corrected views contain no generated content and remain physical RAW images.
\end{itemize}

%% file: sec/2_related_work.tex
\section{Related Work}
\label{sec:related}

\paragraph{Uniform airlight and depth.}
\looseness=1 The atmospheric scattering model of Eq.~\eqref{eq:asm} describes a uniformly lit medium~\cite{koschmieder1924theorie,narasimhan2002vision,lee2016visibility}. Single-image dehazing estimates its airlight and a transmission map, for example with the dark channel prior~\cite{he2011single}, and SmokeGS-R inverts Eq.~\eqref{eq:asm} with this prior to obtain pseudo-clean targets for reconstruction~\cite{smokegsr}. Learning-based dehazing inherits the model through its training data: real hazy and clean pairs require haze machines~\cite{ancuti2018ihaze,ancuti2019dense,ancuti2020nhhaze,zhang2024lmhaze}, so networks mostly learn from pairs synthesized with Eq.~\eqref{eq:asm}~\cite{li2019benchmarking,shin2025hazeflow}, with added degradations~\cite{wu2023ridcp}, or with diffusion-generated haze~\cite{wang2025learning}. Medium-aware reconstruction fits Eq.~\eqref{eq:asm}, or its underwater counterpart~\cite{akkaynak2018revised}, together with a radiance field or a set of Gaussians~\cite{seathrunerf,liu2025i2nerf,dehazegs2025,watersplatting,seasplat,restorgs,tang2024neural}, so the haze of a pixel grows with the depth along its ray, and SmokeSeer, which adds thermal images, regularizes its smoke toward a single color~\cite{smokeseer2025}. Measuring real indoor smoke in linear RAW, we find that this assumption does not hold.

\paragraph{Illumination-dependent scattered light.}
\looseness=1 Where artificial sources light the medium, prior work lets the scattered light depend on the illumination~\cite{narasimhan2003shedding,sharma2021nighttime}. Nighttime dehazing estimates a spatially varying atmospheric light from the hazy image~\cite{li2015nighttime,zhang2017fast,ancuti2020day}; its data synthesis lights scene and haze with the same virtual sources~\cite{zhang2020nighttime} or weights the haze by an illumination map~\cite{liu2026clear}, and SynFog path-traces fog under street lamps and car lights~\cite{xie2024synfog}. In 3D, ScatterNeRF and DehazeNeRF let the airlight vary in space~\cite{ramazzina2023scatternerf,chen2023dehazenerf}. Under water, too, the scattered light is proportional to the ambient light~\cite{akkaynak2018revised}; in murky water lit by lamps beside the camera, it is governed by the lamps rather than by scene depth~\cite{treibitz2009active,tsiotsios2014backscatter,she2025relative}. Visibility through haze and fire smoke likewise depends on how brightly the medium is lit~\cite{duntley1948reduction,jin1997studies}. We find a similar regularity in indoor smoke: the scattered light follows the light sources in distribution and color, and the light ratio sets how much light must be removed. Our simulator builds on this measured structure and, like camera-pipeline models~\cite{zamir2020cycleisp,brooks2019unprocessing}, renders each pair as camera RAW.

\paragraph{Smoke removal in sRGB.}
\looseness=1 Whatever their model of the scattered light, dehazing methods mostly operate on the sRGB images that the camera ISP renders, and they output pixels~\cite{song2023vision,conde2024instructir,jin2025smokebench,ge2026clip,ancuti2026ntire}. For reconstruction through smoke, the common baseline restores every view with a 2D dehazing network~\cite{psd,promptir,moceir,mbtaylorformer,convir} and reconstructs the results with 3D Gaussian splatting~\cite{kerbl3Dgaussians,wu2026mllmgs,ntire2026smoke}; the network regresses every pixel of the restored view. The strongest RealX3D pipelines go one step further and pass each view through a commercial generative image model~\cite{gensmokegs,smokegs,dehazethensplat,wu2026mllmgs}. Their authors report hallucinated detail and structural differences between views~\cite{dehazethensplat,gensmokegs}, so the reconstruction may show content that the camera never recorded. All of these pipelines inherit the limits of their input: in the ISP output, pixel values are no longer proportional to the light.

\paragraph{Restoration in linear RAW.}
Linear RAW keeps this proportionality. Reconstruction in this domain averages zero-mean sensor noise across views and preserves highlights~\cite{mildenhall2022rawnerf,raw3dgs,le3d,pearl2022nan}, and defogging has been combined with demosaicking on the RAW mosaic~\cite{jointdefog}. Parametric enhancement, in turn, avoids predicting pixels~\cite{cui2022iat}: a network estimates only tone curves~\cite{guo2020zerodce,luminancegs2025,cui2024iac}, lookup tables~\cite{zeng2020lut3d}, or bilateral grids~\cite{bilarf,gharbi2017deep,wang2019underexposed,rawild}, which a fixed operator applies. FujinSplat combines the two for smoke: in linear RAW, a network predicts a low-resolution gain field and one transmission, and fixed operators rescale the recorded signal. No content is generated, and Gaussian splatting averages out the amplified sensor noise while fusing the views.

%% file: sec/3_method.tex
\section{Method}
\label{sec:method}

\begin{figure*}[t]
    \centering
    \includegraphics[width=\textwidth]{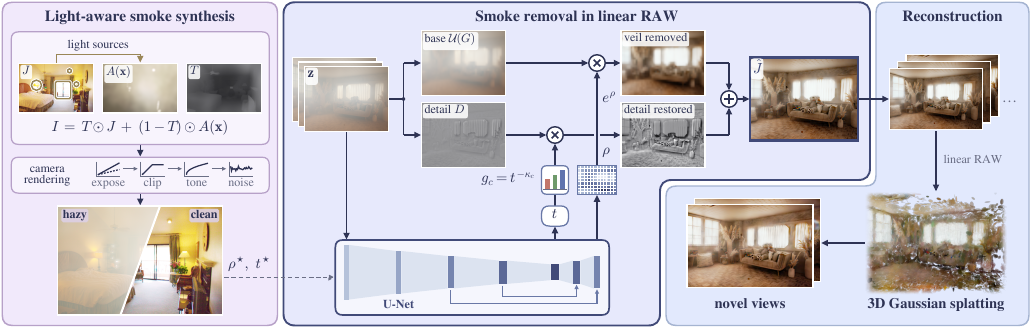}
    \caption{\textbf{Overview of FujinSplat.} \emph{Light-aware smoke synthesis}: a clean image $J$ receives a veil whose light $A(\mathbf{x})$ follows its light sources and is rendered as camera RAW; the labels $\rho^{\star}$ and $t^{\star}$ supervise the network along the dashed line. \emph{Smoke removal in linear RAW}: for a view $\mathbf{z}$, a U-Net predicts per-block log-gains $\rho$ and one transmission $t$; the gains $e^{\rho}$ remove the veil from the base, and the channel gains $g_c=t^{-\kappa_c}$ restore the detail. \emph{Reconstruction}: 3D Gaussian splatting fuses the corrected views in linear RAW.}
    \label{fig:method_pipeline}
\end{figure*}

FujinSplat is a RAW-domain Gaussian splatting framework: it removes the smoke from each RAW view with an ISP-style operator and reconstructs the scene from the corrected views (Fig.~\ref{fig:method_pipeline}).

\subsection{Light-Driven Formation of Indoor Smoke}
\label{sec:formation}

A RAW pixel value is proportional to the light that reaches the sensor, up to the white level. Smoke changes this light in two ways: it attenuates the light from the scene, and it adds the light that it scatters toward the camera. Because light adds, a hazy RAW observation $I$ of a scene with clean radiance $J$ decomposes per pixel as
\begin{equation}
    I_c(\mathbf{x}) = T_c\,J_c(\mathbf{x}) + V_c(\mathbf{x}),
    \quad V_c(\mathbf{x}) = (1-T_c)\,A_c(\mathbf{x}),
    \label{eq:formation}
\end{equation}
where $c$ indexes the color channels, $T_c = T^{\kappa_c}$ with one transmission $T$ for the view and the relative optical thickness $\kappa_c$ of channel $c$, $A$ is the light incident on the smoke along the viewing ray, and $V$ is the veil. Eq.~\eqref{eq:asm} places the spatial variation of the smoke in its transmission, Eq.~\eqref{eq:formation} in the light on it. We test both on the $168$ training poses of RealX3D~\cite{realx3d} that provide paired hazy and clean RAW captures under unchanged lighting.

\begin{figure}[t]
    \centering
    \includegraphics[width=\linewidth]{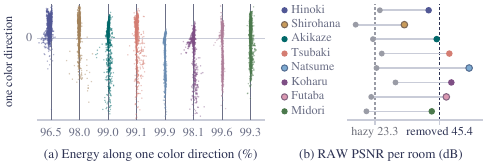}
    \caption{\textbf{The smoke changes every view along one color direction.} (a) The residual $I-J$ of the $168$ paired views in linear RAW, one strip per room in the colors of (b). The vertical axis is its component along the leading color direction, and the horizontal axis the component across it on the same scale. Numbers give the share of the residual energy along this direction. (b) Removing this direction alone, with the amount taken from the clean capture, raises the RAW PSNR of the hazy captures from $23.3$ to $45.4$\,dB on average, marked by the dashed lines.}
    \label{fig:rank_one}
\end{figure}

\paragraph{One transmission and a veil of one color.}
We fit both to every view on a $6\times9$ grid: Eq.~\eqref{eq:formation}, with one transmission per view and a spatially varying veil, and Eq.~\eqref{eq:asm}, with one airlight per view and a spatially varying transmission. Eq.~\eqref{eq:formation} fits all $168$ views better, by a median of $6.1$\,dB. If the light on the smoke also has a single color over the view, the residual $I-J=(1-T_c)\,(A_c-J_c)$ concentrates along one color direction wherever the veil outweighs the attenuated scene. A singular value decomposition of the residual over the pixels of each view shows that it does (Fig.~\ref{fig:rank_one}): the leading color direction holds $98.7\%$ of its energy on average and over $95\%$ in every view.

\paragraph{The veil follows the light sources.}
When we split the veil into its light and opacity factors, the light accounts for $78\%$ of its spatial variation; the rest comes from the variation of $T$ with the path length, which Eq.~\eqref{eq:formation} neglects. The veil rises around light sources, broadly around windows and strongly but locally around lamps. It also takes their color once the spectral dependence of the smoke is accounted for. Attenuation is stronger at short wavelengths, with $\kappa=(1, 1.24, 1.52)$ for the red, green, and blue channels, as also observed for fire smoke~\cite{dobbins1994comparison}; filtered by this spectral opacity $1-T_c$, the color of the light sources measured on the clean captures lies within a median of $1.1^\circ$ of the leading direction of the residual.

\begin{figure}[t]
    \centering
    \includegraphics[width=\linewidth]{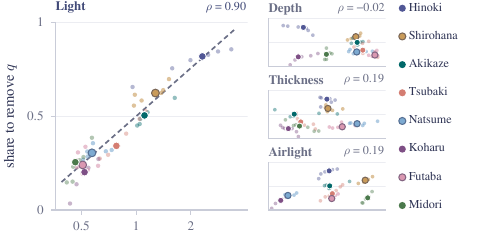}
    \caption{\textbf{The light sets how much to remove.} Share to remove $q$ against the light ratio $\ell$, the scene depth, the optical thickness $\tau$ of the smoke, and the airlight $A_0/W$, the light on the smoke relative to the white level. Large dots are room medians, small dots six views per room; $\rho$ is the Spearman correlation across the eight rooms, and the dashed line a least-squares fit.}
    \label{fig:formation}
\end{figure}

\paragraph{The amount to remove follows the light ratio.}
The first two regularities describe the form of the veil; the third concerns its amount. The light that must be removed from a view is the share $q = 1-\operatorname{med}(Y_J)/\operatorname{med}(Y_I)$ of its luminance $Y$. We compare it with how brightly the smoke is lit relative to the scene, a quantity we call the light ratio, $\ell = A_0/E_{95}$: the median light on the smoke divided by the $95$th percentile of the clean luminance outside highlights. The share follows the light ratio, with Spearman correlations of $0.90$ across the eight rooms and $0.92$ across the $168$ views, against $0.19$ for the optical thickness $\tau=-\ln T$ of the smoke (Fig.~\ref{fig:formation}). The light ratio is largely a property of the room: $84\%$ of its variance across views lies between rooms, and the room medians range from $0.46$ to $2.31$.

\subsection{Light-Aware Indoor Smoke Simulation}
\label{sec:simulation}

\begin{figure}[t]
    \centering
    \includegraphics[width=\linewidth]{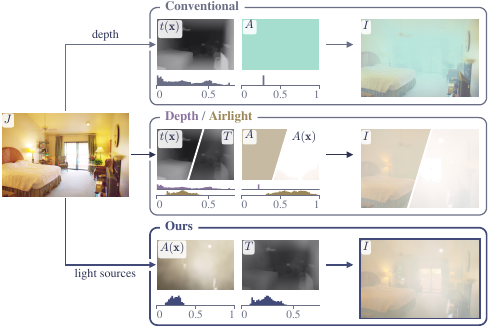}
    \caption{\textbf{Synthesis of training pairs} in the four simulators of Table~\ref{tab:ablation_sim}. Each row composes a hazy image $I$ from the clean image $J$, a transmission, and the light on the smoke; the histograms give both over the image, the light relative to the white level. Conventional synthesis~\cite{li2019benchmarking} and Depth take $t(\mathbf{x})$ from depth and use one airlight $A$. Airlight draws the level of $A(\mathbf{x})$ from the conventional airlight range; the middle row shows Depth left of each diagonal and Airlight right of it. Ours shapes $A(\mathbf{x})$ by the light sources and ties its level to the light of the scene.}
    \label{fig:simulation}
\end{figure}

Paired real captures of indoor smoke are scarce: RealX3D offers eight rooms, which we use only for measurement and evaluation, so the training pairs are simulated. Common dehazing datasets synthesize such pairs by applying Eq.~\eqref{eq:asm} to clean images~\cite{li2019benchmarking}. Measured against Sec.~\ref{sec:formation}, this has two flaws: (1) the veil grows with scene depth instead of following the light sources, and (2) the airlight is drawn from a fixed range relative to white, regardless of the light in the scene (Fig.~\ref{fig:simulation}). We instead synthesize Eq.~\eqref{eq:formation} with the forms measured in Sec.~\ref{sec:formation}. We compose Eq.~\eqref{eq:formation} in linear light and render each pair as camera RAW, the domain in which the network operates. Training is cross-validated by room: the statistics that drive the simulator are taken only from the rooms other than the one under evaluation, so the evaluated room never enters training.

We render smoke over clean linear images from MIT-Adobe FiveK~\cite{bychkovsky2011learning}, ETH3D~\cite{schops2017multi}, and Poly Haven~\cite{polyhaven}, in which windows, lamps, and specular highlights are detected and keep a brightness above the white level. The light on the smoke is $A_c(\mathbf{x}) = A_0\,a(\mathbf{x})\,u_c(\mathbf{x})$. Its shape $a$ rises around the light sources, broadly around windows and narrowly around lamps, as measured in Sec.~\ref{sec:formation}. Its color $u$ is drawn around the measured light colors. Its level $A_0 = w\,\bar{L}$ scales the mean light $\bar{L}$ of the clean image by a random factor $w$ with $\log_2 w \sim \mathcal{N}(0, 0.35^2)$, so the light ratio varies with the scene. Each pair draws a whole-image transmission from the distribution measured on real views, varies it over the image with the path length from monocular depth~\cite{yang2024depthv2}, and applies $T_c=T^{\kappa_c}$. We then expose each pair like a real view, clip it at the camera's white level, and during training add noise of the signal-dependent form measured on real hazy RAW captures. We render $1{,}500$ pairs, split by content into training and validation sets. The supplementary material describes the simulation in detail.

\begin{figure*}[!t]
    \centering
    \includegraphics[width=\textwidth]{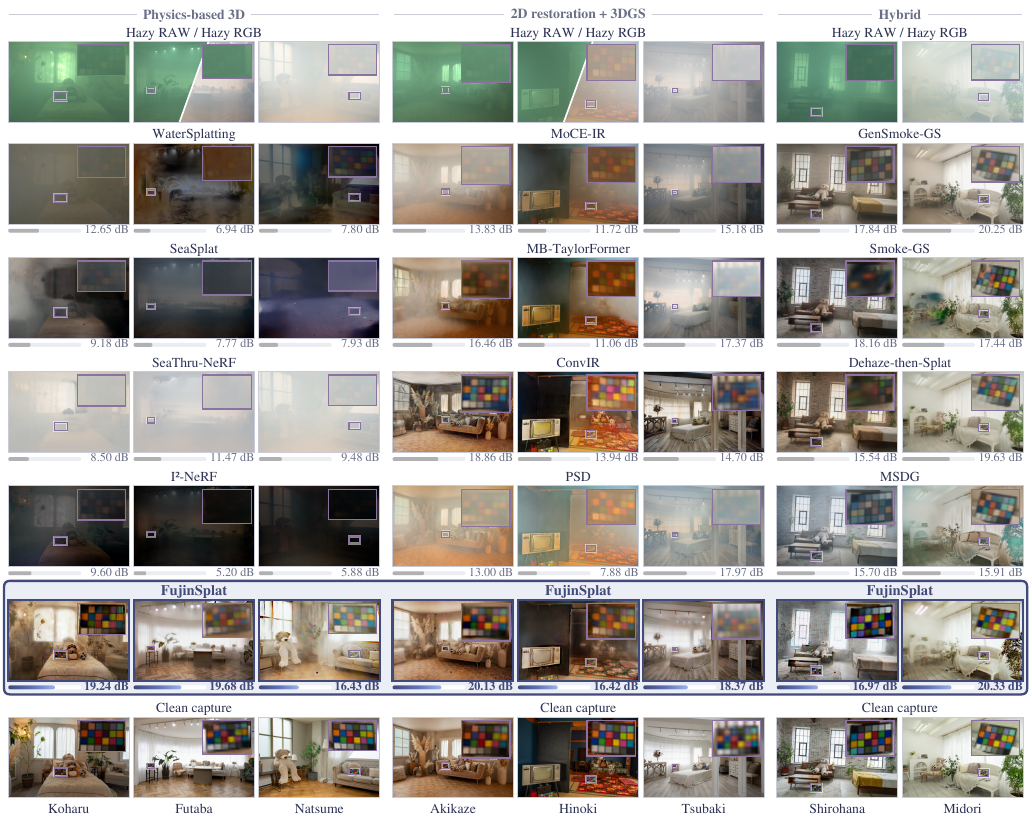}
    \caption{\textbf{Qualitative comparison on RealX3D novel views.} Each group of columns compares FujinSplat with one family of methods; the restoration networks use their official weights and our 3DGS pipeline. Novel views are recorded only in clean air, so the top row shows the nearest hazy training view, as camera RAW left of the diagonal and as the camera's sRGB image right of it. Boxes mark the color chart, enlarged in the corner; bars give the PSNR of each view.}
    \label{fig:qualitative}
\end{figure*}

\subsection{Smoke Removal in Linear RAW}
\label{sec:removal}

Sec.~\ref{sec:formation} fixes what the removal must do in linear RAW: (1) undo a smooth veil that varies over the image, (2) restore the detail that one transmission attenuates, with one gain per channel set by $\kappa$, and (3) return a linear RAW image, the light that the camera would have recorded without smoke. An operator that only rescales components of the observation does all three. A network trained only on the simulated pairs of Sec.~\ref{sec:simulation} predicts its parameters.

\paragraph{Operator.}
Let $\mathbf{z}$ be a linear RAW view, white-balanced and scaled so that the median of its unclipped luminance is $0.18$. We split it into a low-frequency base $\mathcal{U}(G)$, where $G$ holds the means of $8\times8$-pixel blocks and $\mathcal{U}$ is bilinear upsampling, and a detail layer $D=\mathbf{z}-\mathcal{U}(G)$. Under Eq.~\eqref{eq:formation} with a smooth veil, the smoke acts differently on the two layers: the veil falls into the base, and the detail is only attenuated by $T_c$. A network therefore predicts per-block log-gains $\rho$ for the three channels, a low-resolution gain field, and one transmission $t$ for the view, and the output is
\begin{equation}
    \hat{J} = \mathcal{U}\bigl(G\odot e^{\rho}\bigr) + D + \sum_{b} w_b\,(\mathbf{g}-1)\odot D_b,
    \label{eq:operator}
\end{equation}
where $D_b$ are frequency bands of $D$. The base gain removes the veil and restores the level and color of each block. The detail gain $\mathbf{g}$, with $g_c = t^{-\kappa_c}$, undoes the attenuation of the detail with the measured spectral ratio. The weight $w_b$ withholds amplification where the local energy of band $b$ approaches the camera's known noise variance; the supplementary material gives its definition.

\paragraph{A physical output.}
Every term of Eq.~\eqref{eq:operator} rescales a component of the observation, so the linearity of RAW carries over to the output. The output $\hat{J}$ is again a linear RAW image that contains only what the camera recorded. The predicted quantities are physical, $\rho$ a log-ratio of clean to hazy light and $t$ the transmission $T$ of Eq.~\eqref{eq:formation} for the view, so both can be supervised with simulated labels.

\paragraph{Predictor.}
A U-Net predicts $\rho$ from its features at $1/8$ resolution, one value per block and channel, and $t$ from the mean and variance of its features at $1/32$ resolution. Its encoder follows ConvIR~\cite{convir}, and the network has $1.8$M parameters. It sees only the normalized view $\mathbf{z}$, so its predictions do not depend on the exposure of the capture.

\paragraph{Supervision.}
We train with
\begin{equation}
    \mathcal{L} = \bigl\|\Gamma(\hat{J})-\Gamma(J)\bigr\|_1
    + \lambda_\rho\bigl\|\rho-\rho^{\star}\bigr\|_1
    + \bigl|\log t-\log t^{\star}\bigr|,
    \label{eq:loss}
\end{equation}
where $\Gamma$ maps linear RAW to the camera's display encoding, $\rho^{\star}$ is the log-ratio of clean to hazy block means, $t^{\star}$ is the simulated transmission, and $\lambda_\rho=0.2$. The image term passes no gradient to $t$, which is learned from its label alone. This label is necessary: when $t$ is trained through the image term instead, it settles near $0.5$ while its true value is $0.23$, so the detail gain stays below half of its physical value. A pixel loss thus prefers soft output to physically correct amplification; supervising $t$ directly restores texture contrast, as the supplementary material shows.

\paragraph{Reconstruction.}
We reconstruct each scene from its corrected training views with standard 3D Gaussian splatting~\cite{kerbl3Dgaussians}. The Gaussians hold linear RAW color, as in RawNeRF~\cite{mildenhall2022rawnerf}, and the photometric loss, $0.8\,\mathcal{L}_1+0.2\,\mathcal{L}_{\text{D-SSIM}}$, is computed after a fixed tone curve. Averaging across views suppresses the zero-mean noise that the detail gain amplified.

%% file: sec/4_experiments.tex
\section{Experiments}
\label{sec:experiments}

We compare FujinSplat with existing methods on real indoor smoke (Sec.~\ref{sec:main_comparison}), change its training data, its operator, and its input one at a time (Sec.~\ref{sec:ablations}), and vary its remaining settings (Sec.~\ref{sec:sensitivity}).

\begin{table*}[t]
    \centering
    \caption{Novel-view results on the eight RealX3D smoke scenes; each scene value averages four novel views. Colors denote \colorbox{fst}{first}, \colorbox{sed}{second}, and \colorbox{thd}{third} place for every scene and metric.}
    \label{tab:main_results}
    \fontsize{3.0}{3.3}\selectfont
    \setlength{\tabcolsep}{2.161pt}
    \renewcommand{\arraystretch}{0.82}
    \begingroup
    \setlength{\arrayrulewidth}{0.20pt}
    \renewcommand{\colorbox}[2]{\cellcolor{#1}#2}
    \newcommand{\metricrowstrut}{\rule[-0.64ex]{0pt}{2.36ex}}
    \newcommand{\metricHeader}{%
        \makebox[4.1em][l]{\bfseries Metrics}}
    \newcommand{\metricPSNR}{%
        \metricrowstrut
        \makebox[4.1em][l]{PSNR\hfill\scalebox{0.55}{$\uparrow$}}}
    \newcommand{\metricSSIM}{%
        \metricrowstrut
        \makebox[4.1em][l]{SSIM\hfill\scalebox{0.55}{$\uparrow$}}}
    \newcommand{\metricLPIPS}{%
        \metricrowstrut
        \makebox[4.1em][l]{LPIPS\hfill\scalebox{0.55}{$\downarrow$}}}
    \resizebox{\textwidth}{!}{%
    \begin{tabular}{c|l|c|*{9}{r}}
        \hline
        & \rule{0pt}{2.2ex}\textbf{Methods} & \metricHeader
        & \textbf{Aki.} & \textbf{Fut.} & \textbf{Hin.}
        & \textbf{Koh.} & \textbf{Mid.} & \textbf{Nat.}
        & \textbf{Shi.} & \textbf{Tsu.} & \textbf{Avg.} \\
        \hline
        \multirow[c]{12}{*}[-1.2ex]{\makebox[1.5em][c]{%
            \rotatebox[origin=c]{90}{Physics-based 3D Methods}}}
        & & \metricPSNR & 9.54 & 8.64 & 14.08 & 14.23 & 4.64 & 7.81 & 8.22 & 7.24 & 9.30 \\
        & WaterSplatting~\cite{watersplatting} & \metricSSIM & 0.433 & 0.523 & 0.439 & 0.637 & 0.287 & 0.372 & 0.339 & 0.375 & 0.426 \\
        & & \metricLPIPS & 0.757 & 0.677 & 0.822 & 0.650 & 0.779 & 0.693 & 0.706 & 0.638 & 0.715 \\
        \cline{2-12}

        & & \metricPSNR & 11.22 & 8.11 & 13.25 & 9.97 & 5.04 & 8.48 & 9.22 & 8.95 & 9.28 \\
        & SeaSplat~\cite{seasplat} & \metricSSIM & 0.498 & 0.497 & 0.428 & 0.496 & 0.331 & 0.462 & 0.349 & 0.507 & 0.446 \\
        & & \metricLPIPS & 0.861 & 0.727 & 0.673 & 0.622 & 0.850 & 0.865 & 0.846 & 0.637 & 0.760 \\
        \cline{2-12}

        & & \metricPSNR & 8.17 & 10.55 & 4.90 & 7.78 & 12.38 & 8.49 & 8.44 & 11.43 & 9.02 \\
        & SeaThru-NeRF~\cite{seathrunerf} & \metricSSIM & 0.514 & 0.666 & 0.266 & 0.539 & 0.688 & 0.594 & 0.414 & \colorbox{thd}{0.702} & 0.548 \\
        & & \metricLPIPS & 0.672 & 0.620 & 0.736 & 0.653 & 0.575 & 0.614 & 0.739 & 0.597 & 0.651 \\
        \cline{2-12}

        & & \metricPSNR & 7.11 & 5.37 & 13.39 & 9.43 & 3.47 & 6.05 & 5.67 & 6.03 & 7.07 \\
        & I$^2$-NeRF~\cite{liu2025i2nerf} & \metricSSIM & 0.194 & 0.196 & 0.368 & 0.331 & 0.125 & 0.159 & 0.091 & 0.171 & 0.204 \\
        & & \metricLPIPS & 0.606 & 0.578 & 0.543 & 0.458 & 0.626 & 0.591 & 0.638 & 0.580 & 0.578 \\
        \cline{1-12}

        \multirow[c]{15}{*}[-1.0ex]{\makebox[1.5em][c]{%
            \rotatebox[origin=c]{90}{2D Restoration + 3DGS}}}
        & & \metricPSNR & 15.74 & 11.78 & \colorbox{thd}{15.32} & 17.44 & 14.32 & 14.36 & \colorbox{thd}{11.70} & 11.84 & 14.06 \\
        & PromptIR~\cite{promptir} & \metricSSIM & 0.619 & 0.600 & \colorbox{thd}{0.501} & \colorbox{thd}{0.713} & 0.701 & 0.587 & 0.397 & 0.581 & 0.587 \\
        & & \metricLPIPS & 0.385 & 0.384 & \colorbox{thd}{0.475} & \colorbox{thd}{0.297} & 0.369 & 0.428 & 0.549 & 0.360 & 0.406 \\
        \cline{2-12}

        & & \metricPSNR & 14.47 & \colorbox{sed}{15.71} & 13.51 & \colorbox{thd}{18.28} & \colorbox{thd}{16.67} & 12.37 & 10.45 & 16.13 & 14.70 \\
        & MoCE-IR~\cite{moceir} & \metricSSIM & 0.590 & 0.652 & 0.464 & 0.709 & 0.692 & 0.560 & 0.372 & 0.667 & 0.588 \\
        & & \metricLPIPS & 0.409 & 0.401 & 0.486 & 0.300 & 0.382 & 0.440 & 0.555 & 0.361 & 0.417 \\
        \cline{2-12}

        & & \metricPSNR & \colorbox{thd}{16.57} & \colorbox{thd}{15.18} & 13.47 & 16.46 & \colorbox{sed}{17.82} & \colorbox{sed}{15.66} & 10.72 & \colorbox{thd}{17.05} & \colorbox{thd}{15.37} \\
        & MB-TaylorFormer~\cite{mbtaylorformer} & \metricSSIM & 0.616 & 0.650 & 0.493 & 0.676 & \colorbox{sed}{0.728} & \colorbox{thd}{0.608} & 0.387 & 0.678 & \colorbox{thd}{0.605} \\
        & & \metricLPIPS & 0.387 & 0.410 & 0.504 & 0.338 & \colorbox{thd}{0.358} & 0.441 & 0.569 & 0.358 & 0.421 \\
        \cline{2-12}

        & & \metricPSNR & \colorbox{sed}{18.43} & 14.59 & \colorbox{sed}{15.67} & \colorbox{sed}{18.90} & 14.73 & \colorbox{thd}{15.41} & \colorbox{sed}{16.44} & 14.48 & \colorbox{sed}{16.08} \\
        & ConvIR~\cite{convir} & \metricSSIM & \colorbox{sed}{0.767} & \colorbox{sed}{0.736} & \colorbox{sed}{0.593} & \colorbox{sed}{0.759} & 0.690 & \colorbox{sed}{0.704} & \colorbox{sed}{0.710} & 0.621 & \colorbox{sed}{0.697} \\
        & & \metricLPIPS & \colorbox{sed}{0.259} & \colorbox{sed}{0.276} & \colorbox{fst}{\textbf{0.381}} & \colorbox{fst}{\textbf{0.253}} & \colorbox{sed}{0.300} & \colorbox{sed}{0.290} & \colorbox{sed}{0.319} & \colorbox{sed}{0.288} & \colorbox{sed}{0.296} \\
        \cline{2-12}

        & & \metricPSNR & 13.37 & 15.17 & 8.78 & 14.47 & 16.63 & 12.37 & 10.88 & \colorbox{sed}{17.62} & 13.66 \\
        & PSD~\cite{psd} & \metricSSIM & \colorbox{thd}{0.622} & \colorbox{thd}{0.683} & 0.381 & 0.702 & \colorbox{thd}{0.722} & 0.595 & \colorbox{thd}{0.425} & \colorbox{sed}{0.710} & \colorbox{thd}{0.605} \\
        & & \metricLPIPS & \colorbox{thd}{0.353} & \colorbox{thd}{0.379} & 0.494 & 0.323 & 0.376 & \colorbox{thd}{0.420} & \colorbox{thd}{0.534} & \colorbox{thd}{0.351} & \colorbox{thd}{0.404} \\
        \cline{1-12}

        \multirow[c]{3}{*}{\smash{\rotatebox[origin=c]{90}{RAW}}}
        & & \metricPSNR & \colorbox{fst}{\textbf{20.84}} & \colorbox{fst}{\textbf{20.05}} & \colorbox{fst}{\textbf{16.95}} & \colorbox{fst}{\textbf{19.13}} & \colorbox{fst}{\textbf{20.70}} & \colorbox{fst}{\textbf{17.29}} & \colorbox{fst}{\textbf{17.74}} & \colorbox{fst}{\textbf{18.32}} & \colorbox{fst}{\textbf{18.88}} \\
        & \textbf{FujinSplat (Ours)} & \metricSSIM & \colorbox{fst}{\textbf{0.786}} & \colorbox{fst}{\textbf{0.830}} & \colorbox{fst}{\textbf{0.611}} & \colorbox{fst}{\textbf{0.779}} & \colorbox{fst}{\textbf{0.839}} & \colorbox{fst}{\textbf{0.760}} & \colorbox{fst}{\textbf{0.733}} & \colorbox{fst}{\textbf{0.732}} & \colorbox{fst}{\textbf{0.759}} \\
        & & \metricLPIPS & \colorbox{fst}{\textbf{0.231}} & \colorbox{fst}{\textbf{0.225}} & \colorbox{sed}{0.409} & \colorbox{sed}{0.257} & \colorbox{fst}{\textbf{0.232}} & \colorbox{fst}{\textbf{0.250}} & \colorbox{fst}{\textbf{0.271}} & \colorbox{fst}{\textbf{0.268}} & \colorbox{fst}{\textbf{0.268}} \\
        \hline
    \end{tabular}
    }
    \endgroup
\end{table*}

\subsection{Experimental Settings}
\label{sec:experimental_setup}

\paragraph{Dataset.}
We evaluate on the eight smoke scenes of RealX3D~\cite{realx3d}, which provide hazy RAW and JPEG captures at the training poses, clean captures at four held-out novel views per scene, and paired clean RAW captures at the $168$ training poses measured in Sec.~\ref{sec:formation}. Every method reconstructs a scene from its hazy training views only, and the clean novel views serve for evaluation alone. FujinSplat takes the hazy RAW captures, and every baseline takes the input of its native pipeline.

\paragraph{Baselines.}
We compare with three groups of methods. Physics-based 3D methods model the medium inside the reconstruction~\cite{watersplatting,seasplat,seathrunerf,liu2025i2nerf}. Restoration-based pipelines apply 2D restoration networks with their official weights~\cite{promptir,moceir,mbtaylorformer,convir,psd} to every training view and reconstruct the results with the same 3D Gaussian splatting pipeline as ours~\cite{kerbl3Dgaussians}. Hybrid pipelines combine 3D reconstruction with generative or multimodal-language priors~\cite{ntire2026smoke,gensmokegs,smokegs,dehazethensplat,wu2026mllmgs}.

\paragraph{Metrics.}
We report PSNR, SSIM, and LPIPS~\cite{zhang2018unreasonable} between rendered novel views and the official clean references, averaged over the four novel views of each scene and then over the eight scenes. All images are downsampled by $4\times$ with area averaging and quantized to 8 bits before scoring, and LPIPS uses VGG features. We develop the linear RAW output of FujinSplat with an ISP fitted per scene from the hazy RAW captures to the hazy camera JPEGs.

\paragraph{Implementation details.}
Each simulation fold contains $1{,}500$ pairs rendered as described in Sec.~\ref{sec:simulation} from the rooms other than the evaluated one. The network is trained for $10{,}000$ iterations; the supplementary material lists the training settings. We keep the checkpoint that performs best on the simulated validation pairs. Gaussian splatting runs for $18$k iterations from the benchmark's COLMAP points, and we report the mean over three random seeds.

\subsection{Comparison with State-of-the-Art Methods}
\label{sec:main_comparison}

\paragraph{Physics-based and restoration methods.}
Table~\ref{tab:main_results} and Fig.~\ref{fig:qualitative} compare FujinSplat with the physics-based 3D methods and the restoration-based pipelines. FujinSplat reaches an average PSNR of $18.88$\,dB, $2.80$\,dB above ConvIR~\cite{convir}, the strongest baseline, and ranks first in PSNR on all eight scenes. It also obtains the best average SSIM ($0.759$) and LPIPS ($0.268$), ranking first in SSIM on all eight scenes and in LPIPS on six. The physics-based methods model a haze that grows with the depth along each ray. They darken the rooms or leave them veiled, and their average PSNR stays below $9.5$\,dB. Among the restoration networks, MoCE-IR, MB-TaylorFormer, and PSD leave a tinted veil. ConvIR removes most of the veil but shifts the colors and darkens some rooms. In contrast, FujinSplat comes closest to the chart colors of the clean captures.

\paragraph{Comparison with hybrid pipelines.}
Table~\ref{tab:ntire_results} compares FujinSplat with the hybrid pipelines. MLLM-GS~\cite{wu2026mllmgs} supervises its correction with three paired clean captures per scene, so we do not list its scores. Three of the four pipelines reach a higher PSNR than FujinSplat, GenSmoke-GS~\cite{gensmokegs} by $2.07$\,dB. All three pass every training view through a commercial image model. FujinSplat generates no content. Its SSIM of $0.755$ matches the best pipeline, at $0.754$, and its LPIPS of $0.273$ is the lowest.

\begin{table}[!t]
    \centering
    \caption{Comparison with hybrid pipelines on the seven scenes of the NTIRE 2026 RealX3D smoke track~\cite{ntire2026smoke}, all but \textsc{Akikaze}. We score the novel views submitted to the challenge as in Table~\ref{tab:main_results}.}
    \label{tab:ntire_results}
    \footnotesize
    \setlength{\tabcolsep}{2pt}
    \renewcommand{\colorbox}[2]{\cellcolor{#1}#2}
    \begin{tabular*}{\linewidth}{@{\extracolsep{\fill}}lcccc}
        \toprule
        Method & PSNR $\uparrow$ & SSIM $\uparrow$ & LPIPS $\downarrow$ & Generative prior \\
        \midrule
        GenSmoke-GS~\cite{gensmokegs}
            & \colorbox{fst}{\textbf{20.67}} & \colorbox{sed}{0.754} & \colorbox{sed}{0.290} & \gptmodel{1.5} \\
        Smoke-GS~\cite{smokegs}
            & \colorbox{thd}{18.82} & \colorbox{thd}{0.690} & 0.358 & \nanobananamodel{Pro} \\
        Dehaze-then-Splat~\cite{dehazethensplat}
            & \colorbox{sed}{18.98} & 0.597 & 0.342 & \nanobananamodel{Pro} \\
        MSDG~\cite{ntire2026smoke}
            & 17.79 & 0.684 & \colorbox{thd}{0.327} & Difix3D+~\cite{wu2025difix3d} \\
        \midrule
        \textbf{FujinSplat}
            & 18.60 & \colorbox{fst}{\textbf{0.755}} & \colorbox{fst}{\textbf{0.273}} & \nomark \\
        \bottomrule
    \end{tabular*}
\end{table}

\subsection{Ablation Studies}
\label{sec:ablations}

FujinSplat differs from the restoration pipelines of Table~\ref{tab:main_results} in its training data, its operator, and its input. Table~\ref{tab:ablation} changes one of them at a time and retrains the network.

\paragraph{Simulator.}
The simulator rows of Table~\ref{tab:ablation} replace the measured forms of Sec.~\ref{sec:formation} with conventional ones (Fig.~\ref{fig:simulation}). A uniform veil that grows with scene depth, as in Eq.~\eqref{eq:asm}, lowers PSNR to $13.87$\,dB, and the outputs keep a veil around windows and lamps. Drawing the veil level from the conventional airlight range, $0.7$ to $1$ times the display white, regardless of how the scene is lit, lowers it to $15.90$\,dB. Conventional synthesis~\cite{li2019benchmarking} uses both, with a randomly colored airlight composited on $8$-bit display images, and reaches $13.40$\,dB. Trained on it, the network darkens the views of every room by $1.4$ to $1.9$\,EV; trained on our simulator, it darkens them by $0.6$ to $1.3$\,EV. How much a network removes thus depends on the veil levels of its training data, which our simulator ties to the light of each scene.

\paragraph{Operator.}
The operator rows of Table~\ref{tab:ablation} remove parts of Eq.~\eqref{eq:operator}. With the detail gain only, the base receives a single gain per view: PSNR falls from $18.88$ to $15.35$\,dB, and all three metrics degrade in every scene. A single gain cannot remove a veil that varies over the image. Without the base--detail split, one gain map multiplies the whole view and scales the detail down together with the veil. With the base gain only, the detail is left unamplified and stays attenuated. Both variants still correct the low frequencies, which dominate PSNR, and lose less than $1$\,dB, but their SSIM and LPIPS degrade markedly.

\paragraph{RAW input.}
The operator relies on the linear formation that Sec.~\ref{sec:formation} measures in RAW (Fig.~\ref{fig:rank_one}). The last row instead trains and tests the network on the camera JPEG, mapped back to linear values with the camera calibration. SSIM and LPIPS degrade in every scene, to $0.717$ and $0.362$ on average, and PSNR drops by $0.4$\,dB. We attribute this to the camera's processing, which clips highlights and smooths low-contrast detail before the network sees the image. The noise variance of the JPEG is also unknown and must be estimated from the image.

\begin{table}[t]
    \centering
    \caption{Top: ablations of the simulator, the operator of Eq.~\eqref{eq:operator}, and the input, scored as in Table~\ref{tab:main_results}; the variants use one 3DGS seed, and the simulator rows one network each, trained on a single fold. Bottom: sensitivity, with the change of the average novel-view PSNR in dB next to each value and our setting in bold; each setting uses one network and one 3DGS seed.}
    \label{tab:ablation}\label{tab:ablation_sim}\label{tab:ablation_method}\label{tab:sensitivity}
    \footnotesize
    \setlength{\tabcolsep}{2.2pt}
    \begin{tabular*}{\linewidth}{@{}l@{\hspace{5pt}}l@{\extracolsep{\fill}}ccc@{}}
        \toprule
        \multicolumn{2}{@{}l}{Variant} & PSNR$\uparrow$ & SSIM$\uparrow$ & LPIPS$\downarrow$ \\
        \midrule
        \multicolumn{2}{@{}l}{FujinSplat} & 18.88 & 0.759 & 0.268 \\
        \addlinespace[3pt]
        \textit{Simulator} & Veil grows with depth & 13.87 & 0.641 & 0.356 \\
         & Veil level from airlight range & 15.90 & 0.696 & 0.319 \\
         & Conventional synthesis~\cite{li2019benchmarking} & 13.40 & 0.622 & 0.352 \\
        \addlinespace[3pt]
        \textit{Operator} & No base--detail split & 17.94 & 0.606 & 0.519 \\
         & Base gain only & 18.11 & 0.642 & 0.442 \\
         & Detail gain only & 15.35 & 0.674 & 0.328 \\
        \addlinespace[3pt]
        \textit{Input} & Camera JPEG & 18.44 & 0.717 & 0.362 \\
        \midrule
    \end{tabular*}\\[-0.5pt]
    \newcommand{\dl}[1]{\,{\scriptsize\textcolor{black!55}{$#1$}}}
    \setlength{\tabcolsep}{1.5pt}
    \begin{tabular*}{\linewidth}{@{}l@{\extracolsep{\fill}}llll@{}}
        Veil-level jitter (EV) & 0\dl{+0.02} & 0.2\dl{0.00} & \textbf{0.35} & 0.7\dl{-0.09} \\
        Veil chroma ($\times$) & 0\dl{+0.06} & 0.5\dl{-0.01} & \textbf{1} & 1.5\dl{-0.02} \\
        Block size (px) & 4\dl{+0.40} & \textbf{8} & 16\dl{-0.75} & 32\dl{-1.57} \\
        Pairs & 375\dl{-0.01} & 750\dl{-0.12} & \textbf{1{,}500} & 3{,}000\dl{-0.22} \\
        \bottomrule
    \end{tabular*}
\end{table}

\subsection{Sensitivity Analysis}
\label{sec:sensitivity}

The bottom of Table~\ref{tab:sensitivity} varies one setting at a time and evaluates novel views as in Table~\ref{tab:main_results}. The simulator settings change the average PSNR by at most $0.22$\,dB, from no jitter of the veil level to $0.7$\,EV, from a gray veil to $1.5$ times the default chroma, and from $375$ to $3{,}000$ simulated pairs. The block size of the base gain matters more. Blocks of $16$ and $32$\,px lose $0.75$ and $1.57$\,dB, and a single gain per view loses $3.5$\,dB (Table~\ref{tab:ablation}). We attribute this to the local rise of the veil around lamps (Sec.~\ref{sec:formation}), which a coarser gain field cannot follow. Blocks of $4$\,px raise PSNR by $0.40$\,dB but lower SSIM by $0.014$ and raise LPIPS by $0.014$, so we keep $8$\,px blocks.

%% file: sec/5_conclusion.tex
\section{Conclusion}
\label{sec:conclusion}

We have shown that indoor smoke is governed by the light that illuminates it. In linear RAW, it attenuates each view by one transmission and adds a single-colored veil that follows the light sources, and the light ratio sets how much light must be removed. FujinSplat therefore removes the smoke by rescaling what the camera recorded. A network trained only on a light-aware simulator predicts a low-resolution gain field and a supervised transmission, and fixed ISP-style operators apply them. The corrected views contain no generated content. On RealX3D, FujinSplat raises the novel-view PSNR of the strongest non-generative baseline from $16.08$ to $18.88$\,dB.

\paragraph{Limitations.}
The network does not infer the light ratio of a room from its views; it removes about as much veil as its training data suggest (Sec.~\ref{sec:ablations}). Where the signal behind the smoke falls below the sensor noise, the detail stays unrecovered, whereas a generative pipeline would synthesize it. Our simulator models the diffuse light of windows and lamps, not directional light shafts or rapidly changing smoke. We evaluate on eight rooms captured with one camera, since paired real captures of indoor smoke remain scarce.

\section*{Acknowledgment}
This work was partially supported by JST Moonshot R\&D Grant Number JPMJPS2011 and by the New Energy and Industrial Technology Development Organization (NEDO) under project JPNP26007. Ziteng Cui was supported by JSPS KAKENHI Grant Number JP26K21296. Shuhong Liu was supported by JST BOOST, Japan Grant Number JPMJBS2418.

\section*{LLM Usage Statement}
We used LLM-based tools in this work. For implementation and experiments, they wrote and debugged code and ran experiments as agents under the authors' direction. For writing, they helped draft and edit the text and prepare figures. The research ideas, the method and the conclusions are the authors' own. The authors verified all results, references, text and code, and take full responsibility for the content of this paper.

%% file: sec/X_suppl.tex
\clearpage
\setcounter{page}{1}
\renewcommand{\thepage}{S\arabic{page}}
\maketitlesupplementary
\label{supplementary:start}
\raggedbottom
\setcounter{section}{0}
\renewcommand{\thesection}{\Alph{section}}
\renewcommand{\theHsection}{supp.\Alph{section}}
\setcounter{figure}{0}
\renewcommand{\thefigure}{S\arabic{figure}}
\renewcommand{\theHfigure}{supp.\arabic{figure}}
\setcounter{table}{0}
\renewcommand{\thetable}{S\arabic{table}}
\renewcommand{\theHtable}{supp.\arabic{table}}
\setcounter{equation}{0}
\renewcommand{\theequation}{S\arabic{equation}}
\renewcommand{\theHequation}{supp.\arabic{equation}}
\newcommand{\rankcolors}{Colors denote \colorbox{fst}{first},
  \colorbox{sed}{second}, and \colorbox{thd}{third} place for every scene
  and metric. Ranks use unrounded values.}
\newcounter{suppref}
\makeatletter
\DeclareRobustCommand{\citeS}[1]{\def\@citeSsep{}[\@for\@citeSkey:=#1\do{\@citeSsep\ref{sref:\@citeSkey}\def\@citeSsep{, }}]}
\makeatother
\renewcommand{\topfraction}{0.95}
\renewcommand{\dbltopfraction}{0.95}
\renewcommand{\textfraction}{0.12}
\renewcommand{\bottomfraction}{0.6}
\renewcommand{\floatpagefraction}{0.8}
\renewcommand{\dblfloatpagefraction}{0.8}
\setcounter{dbltopnumber}{2}
\makeatletter\setlength{\@fptop}{0pt}\setlength{\@dblfptop}{0pt}\makeatother

\setcounter{topnumber}{1}

\section{Light-Aware Smoke Synthesis}
\label{sec:suppl_simulation}

\begin{figure*}[t]
    \centering
    \includegraphics[width=\textwidth]{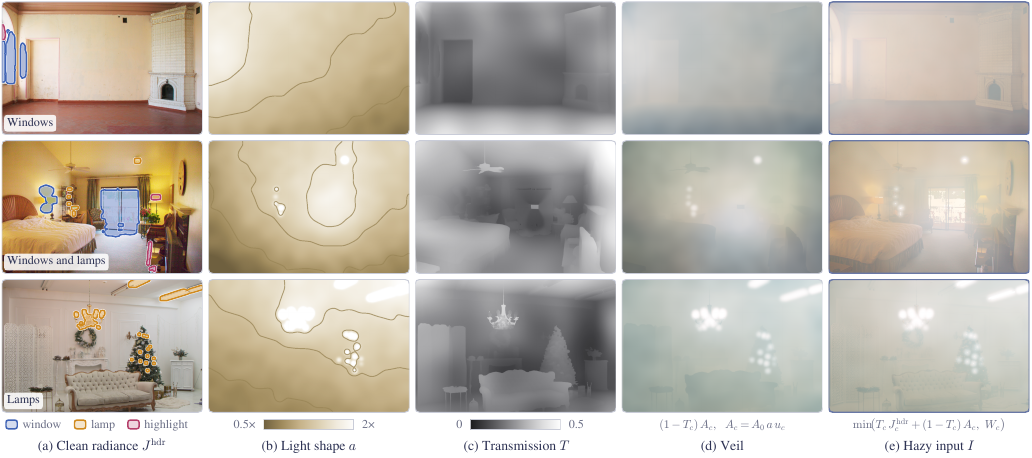}
    \caption{\textbf{Rendering of three training pairs.} (a) The clean radiance $J^{\mathrm{hdr}}$ with the detected windows (blue), lamps (ochre), and specular highlights (pink). (b) The shape $a$ of the light on the smoke with its contours at $0.7$, $1$, and $1.4$. (c) The transmission $T$, shown for luminance. (d) The veil. (e) The hazy input $I$ before noise. The rows hold a pair with detected windows, one with windows and lamps, and one with lamps.}
    \label{fig:suppl_sim_stages}
\end{figure*}

\begin{figure}[!b]
    \centering
    \includegraphics[width=\linewidth]{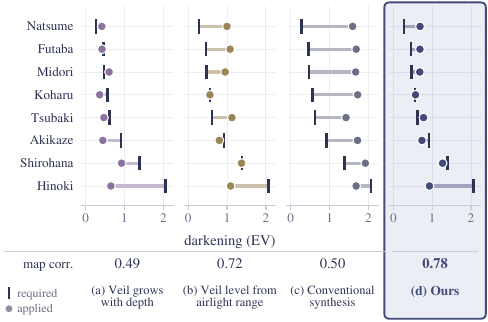}
    \caption{\textbf{Darkening required and applied}, as means over the training poses of a room. Each simulator of Table~\ref{tab:ablation} has one network, trained on the fold of \textsc{Hinoki} with the same settings and applied to all eight rooms. The darkening of a view is the negative mean of its log-gains $\rho$ over the blocks, weighted over the channels by luminance, and the required darkening is that of the labels $\rho^{\star}$ from the paired clean capture. The last row is the correlation between the maps of $\rho$ and $\rho^{\star}$ within a view, averaged over the views of a room and then over the rooms.}
    \label{fig:suppl_darkening}
\end{figure}

The simulation of Sec.~\ref{sec:simulation} of the main paper reproduces the forms measured in Sec.~\ref{sec:formation} and samples its parameters broadly.
Table~\ref{tab:suppl_sim_params} lists the parameters, and Fig.~\ref{fig:suppl_sim_stages} follows three pairs through the steps. Table~\ref{tab:suppl_sim_same_ruler} measures its pairs like the real views, and Fig.~\ref{fig:suppl_darkening} shows how each simulator shapes the network.
Table~\ref{tab:ablation} gives the novel-view PSNR when the shape, the level, or the whole synthesis is replaced, and its lower part the sensitivity to the chroma and the level jitter of the veil.

\paragraph{Light sources.}
The veil rises broadly around windows and strongly but locally around lamps (Sec.~\ref{sec:formation}), so the simulator must tell them apart.
It separates windows, lamps, and specular highlights in each clean image by area, clipping, and contrast with the surroundings.
The level of the veil follows the light of the scene, yet photographs clip lamps and windows.
The detected light sources therefore keep a brightness above the white level $W$ of the simulated camera.
Daylight windows in the photographs are often $20$ to $100$ times brighter than the median, and those of RealX3D $4.5$ to $11$ times.
We therefore compress the window pixels until their mean matches that of a training view.
The result is the radiance $J^{\mathrm{hdr}}$, and the clean target is $J=\min(J^{\mathrm{hdr}},W)$.

\paragraph{Shape.}
Most of the spatial variation of the veil comes from the light on the smoke (Sec.~\ref{sec:formation}).
This light, $A_c(\mathbf{x}) = A_0\,a(\mathbf{x})\,u_c(\mathbf{x})$ in the main paper, has the shape
\begin{equation}
    \log a(\mathbf{x}) = \log\bigl(1 + \alpha\,R(\mathbf{x})\bigr) + v\,y + h\,x + n(\mathbf{x}) - m.
    \label{eq:suppl_shape}
\end{equation}
The response $R$ is obtained by blurring the luminance of each class of light sources with Gaussians and summing the results with non-negative weights that are fitted to the measured light on the smoke.
Windows receive broad kernels, and lamps narrow and medium ones.
The narrow kernel acts on every lamp separately, so that a row of lamps does not merge into one band.
The slopes $v$ and $h$ over the image coordinates $x,y\in[-0.5,0.5]$ reproduce the gradients of the measured light, which grows toward the top.
The term $n$ adds smooth variation, and $m$ sets the median of $a$ outside highlights to $1$.

\begin{table}[t]
    \centering
    \caption{\textbf{Parameters of the simulator.} Training views: drawn from the views of the rooms other than the one under evaluation. Fixed: a constant shared by all pairs. Prior: a broad distribution. Here $\epsilon\sim\mathcal{N}(0,1)$, and s.d.\ is the standard deviation.}
    \label{tab:suppl_sim_params}
    \footnotesize
    \setlength{\tabcolsep}{2.5pt}
    \begin{tabular}{@{}>{\raggedright\arraybackslash}p{0.255\linewidth}>{\raggedright\arraybackslash}p{0.485\linewidth}>{\raggedright\arraybackslash}p{0.205\linewidth}@{}}
        \toprule
        Parameter & Distribution or value & Source \\
        \midrule
        \multicolumn{3}{@{}l}{\textit{Light sources}} \\
        Lamp radiance & log-uniform in $[4,25]\,W$ & prior \\
        Highlight radiance & log-uniform in $[1,2.5]\,W$ & prior \\
        Window brightness & mean of a view $\times\,e^{0.1\epsilon}$, limited to $[3,16]$ times the median & training views \\
        \addlinespace[2pt]
        \multicolumn{3}{@{}l}{\textit{Shape $a$}} \\
        Window kernels & $\sigma = 32$, $64$, $256$\,px & fixed \\
        Lamp kernels & $\sigma = 8$, $32$, $64$\,px; per lamp at $4$\,px & fixed \\
        Highlight kernel & $\sigma = 32$\,px & fixed \\
        Input cap of $R$ & $17.5$ times the median & fixed \\
        Lamp energy cap & 90th percentile & fixed \\
        Strength $\alpha$ & log-normal, median $0.85$, log s.d.\ $0.25$, in $[0.4,2]$ & prior \\
        Slope $v$ & that of a view $+\,0.05\,\epsilon$ & training views \\
        Slope $h$ & $\pm 0.6$ times that of a view & training views \\
        Variation $n$ & $16$--$64$\,px and $>64$\,px, with $0.3$ and $0.4$ times the s.d.\ of a view & training views \\
        \addlinespace[2pt]
        \multicolumn{3}{@{}l}{\textit{Color $u$}} \\
        Log chromaticity & $0.45$ times the gray-world value of the clean image, plus the offset and spread of the views & training views \\
        Local shift & log-normal, median $0.15$, log s.d.\ $0.5$, in $[0.025,0.56]$ & prior \\
        \addlinespace[2pt]
        \multicolumn{3}{@{}l}{\textit{Level and transmission}} \\
        Factor $w$ & as in Sec.~\ref{sec:simulation} & prior \\
        Transmission $T$ & that of a view, redrawn if the veil exceeds $30$ times the direct signal & training views \\
        Range of $\log r$ & that of a view & training views \\
        Gradient $v_\tau$ & that of a view $+\,0.05\,\epsilon$ & training views \\
        Variation $n_\tau$ & $\sigma = 32$\,px, s.d.\ $0.08$ & prior \\
        $\kappa_B/\kappa_R$ & log-normal around $1.52$, log s.d.\ $0.09$, in $[1.30,1.95]$; for $15\%$ of the pairs uniform in $[0.92,1.08]$ & prior \\
        $\kappa_G/\kappa_R$ & $(\kappa_B/\kappa_R)^{0.51}$ & fixed \\
        \addlinespace[2pt]
        \multicolumn{3}{@{}l}{\textit{Camera and noise}} \\
        Exposure, $W$ & those of a view & training views \\
        Noise $a_c$, $b_c$ & those of a room, s.d.\ scaled by $0.682$ & training views \\
        \bottomrule
    \end{tabular}
\end{table}

\paragraph{Color.}
The light on the smoke has one color per view, that of its light sources (Sec.~\ref{sec:formation}).
The color $u$ is therefore drawn around the light colors of the training views, with a weak local shift toward the color of the nearby scene.
All captures of the eight smoke scenes share one camera white balance, under which the light on the smoke lies within a median of $2.4^\circ$ of the neutral axis.
Setting the veil chroma between $0$ and $1.5$ times its default changes the novel-view PSNR by at most $0.06$\,dB (Table~\ref{tab:sensitivity}).

\paragraph{Level.}
The share of light to remove follows the light ratio (Sec.~\ref{sec:formation}), so the level of the light on the smoke must follow the light of the scene.
In $A_0 = w\,\bar{L}$ of Sec.~\ref{sec:simulation}, the mean light $\bar{L}$ is that of $J^{\mathrm{hdr}}$, so that the restored lamps and windows contribute to it.
On RealX3D, the room medians of $A_0/\bar{L}$ lie between $0.67$ and $1.32$ in seven rooms.
In the eighth, the main lamp is clipped in every view, so $\bar{L}$ cannot be measured.

\paragraph{Transmission.}
One transmission per view describes the attenuation, up to a variation with the path length (Sec.~\ref{sec:formation}).
Each pair takes the $T$ of a random training view and varies the optical thickness as
\begin{equation}
    \tau = \tau_0 \exp\bigl(0.95\,(\log r - \operatorname{med}\log r) + v_\tau\,y + n_\tau\bigr),
    \label{eq:suppl_tau}
\end{equation}
where $r$ is the path length from monocular depth~\citeS{yang2024depthv2}, with the range of a training view, $0.95$ is the slope measured in Sec.~\ref{sec:suppl_measurement}, $v_\tau\,y$ is a vertical gradient, and $n_\tau$ is a smooth variation (Table~\ref{tab:suppl_sim_params}).
We choose $\tau_0$ so that the median transmission of the simulated view equals that of the drawn real view, both measured in the same way.
Attenuation is stronger at short wavelengths (Sec.~\ref{sec:formation}), so the channels transmit $T_c = e^{-\kappa_c\tau}$ with ratios of $\kappa$ drawn around the measured ones.

\paragraph{Composition and camera.}
Smoke adds its light before the camera records the image, and the network reads RAW.
We therefore compose each pair in linear light,
\begin{equation}
    I_c = \min\bigl(T_c\,J^{\mathrm{hdr}}_c + (1-T_c)\,A_c,\; W_c\bigr),
\end{equation}
so that the light sources shine through the smoke and clip at the white level, as in the real captures.
The exposure matches the median of the hazy camera JPEG of a random training view, drawn independently of the smoke.

\paragraph{Noise.}
Removing smoke amplifies noise, so we add Gaussian noise with the affine variance of Sec.~\ref{sec:suppl_operator} during training.

\paragraph{Comparison with the real views.}
We measure the real views, the pairs of our simulator, and the pairs of the other three simulators of Table~\ref{tab:ablation} with the estimators of Sec.~\ref{sec:suppl_measurement} (Table~\ref{tab:suppl_sim_same_ruler}).
For our simulator, the light ratio, the share to remove, and the brightening lie within the variation obtained by resampling the eight real rooms with replacement.
With a veil that grows with depth, Eq.~\eqref{eq:formation} fits better than Eq.~\eqref{eq:asm} in $1.3\%$ of the pairs, and the light accounts for almost none of the veil variation.
A level from the airlight range raises $\ell$, $q$, and the brightening above those of the real views.
Conventional synthesis composites on display values, to which the linear models apply only in part.

\begin{table}[t]
    \centering
    \caption{\textbf{The simulators of Table~\ref{tab:ablation}, measured like the real views}, on $300$ pairs per simulator from the fold of \textsc{Hinoki}. The first column is the share of views in which Eq.~\eqref{eq:formation} fits better than Eq.~\eqref{eq:asm}. The other columns are medians: the advantage of Eq.~\eqref{eq:formation} over Eq.~\eqref{eq:asm} in dB, the share of the veil variation due to the light, the light ratio $\ell$, the share to remove $q$, and the brightening $\operatorname{med}(Y_I)/\operatorname{med}(Y_J)$. The brightening is measured on the pixels unclipped in both images of a pair, without the $2$-pixel neighborhoods of clipped pixels, and $q$ on the pixels unclipped in the hazy image.}
    \label{tab:suppl_sim_same_ruler}
    \footnotesize
    \setlength{\tabcolsep}{0pt}
    \begin{tabular*}{\linewidth}{@{}l@{\extracolsep{\fill}}cccccc@{}}
        \toprule
        & Eq.~\eqref{eq:formation} & Adv. & Light & $\ell$ & $q$ & Bright. \\
        \midrule
        Real & $100\%$ & $6.1$ & $0.78$ & $0.68$ & $0.34$ & $1.54$ \\
        Ours & $97\%$ & $7.2$ & $0.88$ & $0.61$ & $0.41$ & $1.75$ \\
        Veil grows with depth & $1.3\%$ & $-5.0$ & $0.01$ & $0.58$ & $0.41$ & $1.76$ \\
        Veil level from airlight range & $97\%$ & $4.7$ & $0.89$ & $1.09$ & $0.63$ & $2.78$ \\
        Conventional synthesis~\citeS{li2019benchmarking} & $9.3\%$ & $-1.6$ & $-0.04$ & $1.20$ & $0.58$ & $2.47$ \\
        \bottomrule
    \end{tabular*}
\end{table}

\begin{figure}[t]
    \centering
    \includegraphics[width=\linewidth]{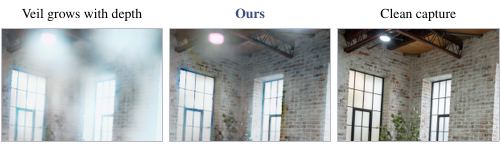}
    \caption{\textbf{Veil left around light sources.} A novel view of \textsc{Shirohana} with a ceiling lamp and windows, from one 3DGS seed. From left to right: the result of the network trained on a veil that grows with depth (Table~\ref{tab:ablation}), our result, and the clean capture. The first leaves a veil around the lamp and the windows.}
    \label{fig:suppl_sim_veil}
\end{figure}

\paragraph{Effect on the network.}
To avoid unnecessary computational cost, each of the other simulators of Table~\ref{tab:ablation} trains one network on the fold of \textsc{Hinoki} and applies it to all eight rooms. The overlap of this fold with the evaluated rooms favors these networks, which fall $3.0$ to $5.5$\,dB below ours in novel-view PSNR.
Fig.~\ref{fig:suppl_darkening} compares how much the network trained on each simulator darkens the views of a room with how much the room requires, and its last row gives the correlation between the maps of $\rho$ and $\rho^{\star}$ within a view.
A veil that grows with depth keeps our level, but its network darkens every room by $0.1$ to $0.3$\,EV less than ours, and the correlation is $0.49$ against $0.78$ for ours, lower in $194$ of the $195$ training poses.
A level from the airlight range keeps our shape. The correlation is $0.72$, and in the five rooms that require the least darkening the network removes $0.47$\,EV more than required on average, against $0.20$\,EV for ours.
Conventional synthesis replaces both. Its correlation is $0.50$, and its network removes $0.5$ to $1.3$\,EV more than required in seven of the eight rooms.
All four networks remove too little in \textsc{Hinoki}.
In the novel view of Fig.~\ref{fig:suppl_sim_veil}, the network trained on a veil that grows with depth leaves a veil around the ceiling lamp and the windows.

\begin{figure*}[t]
    \centering
    \includegraphics[width=\textwidth]{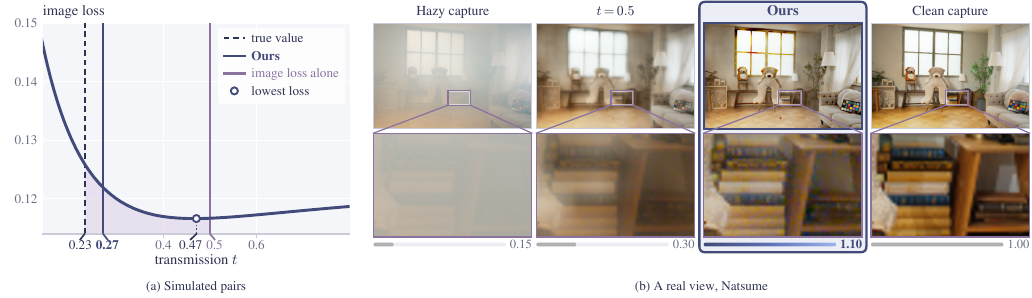}
    \caption{\textbf{Learning $t$ with and without its label.} (a) Image term of Eq.~\eqref{eq:loss} against $t$, averaged over the $84$ simulated validation pairs of the fold of \textsc{Hinoki}, with the base of our network fixed. Ours is the median $t$ of our network on these pairs, and image loss alone marks where $t$ settles when the network learns it through the image term instead. (b) A view of \textsc{Natsume} and a close-up, corrected with $t=0.5$ and by our network. The bars give the amplitude of the $2$--$4$\,px band relative to the clean capture and end at $1$.}
    \label{fig:suppl_t_supervision}
\end{figure*}

\section{Supervising \texorpdfstring{$t$}{t} Restores Texture Contrast}
\label{sec:suppl_t_supervision}

In Sec.~\ref{sec:removal} of the main paper, $t$ is learned from its label alone.
Learned through the image term instead, it settles near $0.5$, while its true value is $0.23$.

We fix the base of our network and scan $t$ alone on the $84$ simulated validation pairs of the fold of \textsc{Hinoki} (Fig.~\ref{fig:suppl_t_supervision}a).
The image term is lowest at $t=0.47$ and rises from $0.117$ to $0.126$ toward the true value, whereas our network, which learns $t$ from its label, predicts a median of $0.27$ on the same pairs.

Fig.~\ref{fig:suppl_t_supervision}b compares $t=0.5$ with our network on a view of \textsc{Natsume}, the room with the densest smoke.
We measure the amplitude of the $2$--$4$\,px band relative to the clean capture, with $1$ denoting equal amplitudes.
Over the $195$ training poses, this ratio averages $1.06$ for our network.
With the same base and $t$ set to $0.5$, it falls to $0.49$ and is lower in all $195$ views.

\section{Training and Implementation Details}
\label{sec:suppl_training}

\paragraph{Network.}
The U-Net has five levels at $1/2$ to $1/32$ resolution, with $16$ to $256$ channels, and its decoder ends at $1/8$ resolution.
The log-gains are bounded as $\rho = 4\tanh(\tilde{\rho}/4)$, where $\tilde{\rho}$ is the raw output of the head.
The transmission is $t = 0.05 + 0.95\,\mathrm{sigmoid}(p + c_0)$, where $p$ is the output of a perceptron with $128$ hidden units and $c_0$ sets $t = 0.98$ at initialization.
The last layers of both heads start at zero.
The network receives $\mathbf{z}$ through the power curve $(\max(\mathbf{z},0)/4)^{1/2.2}$, and the operator acts on the linear $\mathbf{z}$.

\paragraph{Training.}
We train each fold with batch size $8$ on $256\times256$ crops, using Adam~\citeS{kingma2015adam} with a learning rate of $4\times10^{-4}$, $1{,}000$ warm-up iterations, and cosine decay.
At the working resolution of Sec.~\ref{sec:suppl_evaluation}, the predictor takes $4.1$\,ms per view on one NVIDIA A100, and the full correction with the operator takes $13.6$\,ms.

\paragraph{Reconstruction.}
We use the reference implementation of 3D Gaussian splatting~\citeS{kerbl3Dgaussians} with the following changes.
The COLMAP points of the benchmark are mapped into the frame of the official poses by a similarity transform.
Their initial color is the per-channel median of the corrected views.
Densification ends at iteration $6{,}000$, and the opacities are never reset.

\begin{figure}[t]
    \centering
    \includegraphics[width=\linewidth]{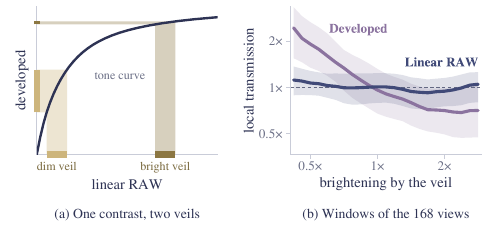}
    \caption{\textbf{Transmission before and after developing.} (a) The same clean contrast under a dim and a bright veil. The two hazy contrasts are equal in RAW and unequal after the camera's tone curve. (b) Local transmission in the $32$-pixel windows of the $168$ views against the brightening by the veil, which is the hazy-to-clean brightness ratio in RAW. Both are relative to the median of the view. Lines are medians, bands quartiles, and the dashed line marks one transmission per view.}
    \label{fig:suppl_apparent_t}
\end{figure}

The Gaussians store camera RAW values.
Their renderings are not clipped, so highlights above the white level keep their linear values.
A loss on linear values is dominated by errors in bright areas~\citeS{mildenhall2022rawnerf}.
We therefore compute the loss of Sec.~\ref{sec:removal} after the fixed tone curve
\begin{equation}
    \mathcal{T}(x) = \bigl(\eta\,\max(x,0) + 10^{-4}\bigr)^{1/2.2} - \bigl(10^{-4}\bigr)^{1/2.2},
    \label{eq:suppl_tone}
\end{equation}
where $\eta$ is $0.18$ divided by the median RAW value of the room's hazy training views.
One reconstruction takes 3 to 5 minutes on one NVIDIA A100.

\section{Evaluation Protocol}
\label{sec:suppl_evaluation}

\paragraph{Working resolution.}
The JPEGs of the benchmark measure between $1734\times1155$ and $1810\times1188$ pixels.
We crop every image at its right and bottom edges to a multiple of 4 pixels and average $4\times4$ blocks.
The working resolution lies between $433\times288$ and $452\times297$ pixels.
The RAW captures are read as linear camera RGB and averaged in the same way.

\paragraph{FujinSplat.}
We correct each hazy RAW training view, reconstruct the scene once per seed, and render its four novel views in camera RAW.
We develop the linear RAW output of FujinSplat with an ISP fitted per scene from the hazy RAW captures to the hazy camera JPEGs.
We clip the camera RAW values to $[0,1]$, apply the camera white balance, and then use three channel gains, a $3\times3$ color matrix, and one monotone tone curve per output channel.
Each curve interpolates $129$ knots linearly on the axis $x/(x+0.2)$ for nonnegative linear intensity $x$.
We initialize the matrix from the camera metadata and sample $8{,}192$ corresponding pixel pairs per hazy training view at the working resolution.
We fit the ISP for $2{,}000$ Adam steps with $8{,}192$ pairs per batch, minimizing mean squared error in display RGB with a smoothness penalty on the tone curves.
The fitted parameters are shared across the pixels and views of the scene.
The developed image is clipped to $[0,1]$ before the quantization.

\paragraph{Restoration baselines.}
Each network restores the hazy camera JPEG of every training view at the resolution of the benchmark.
The restored images are averaged to the working resolution, and the scene is reconstructed from them as in Sec.~\ref{sec:suppl_training}.
The Gaussians then hold RGB values, their renderings are clipped to $[0,1]$, and the loss compares RGB values without the tone curve of Eq.~\eqref{eq:suppl_tone}.

\begin{table}[t]
    \centering
    \caption{\textbf{The measurements of Sec.~\ref{sec:formation} in three domains}, on the same views and with the same procedure. Developed: both views of a pair after the camera's per-channel tone curve at a shared exposure. JPEG: the JPEGs written by the camera. Local transmission: the transmission in the windows of Fig.~\ref{fig:suppl_apparent_t}b. Slope: the slope of its logarithm against the log hazy-to-clean brightness ratio of the domain.}
    \label{tab:suppl_domains}
    \footnotesize
    \setlength{\tabcolsep}{0pt}
    \begin{tabular*}{\linewidth}{@{}l@{\extracolsep{\fill}}ccc@{}}
        \toprule
        Measurement & RAW & Developed & JPEG \\
        \midrule
        Local transmission, slope & $-0.09$ & $-0.76$ & $-0.81$ \\
        Eq.~\eqref{eq:formation} over Eq.~\eqref{eq:asm}, median (dB) & $+6.1$ & $+7.0$ & $+4.8$ \\
        Views where Eq.~\eqref{eq:formation} fits better & $168/168$ & $168/168$ & $168/168$ \\
        Spearman of $q$ and $\ell$, rooms / views & $0.90$ / $0.92$ & $0.33$ / $0.55$ & $0.52$ / $0.63$ \\
        Range of $\ell$ across rooms & $5.0\times$ & $1.6\times$ & $1.6\times$ \\
        \bottomrule
    \end{tabular*}
\end{table}

\paragraph{Metrics.}
PSNR and SSIM are computed on the 8-bit images divided by 255, with a data range of 1.
SSIM uses the implementation of scikit-image with its default $7\times7$ uniform window, averaged over the three channels.
LPIPS~\citeS{zhang2018unreasonable} uses VGG features on images scaled to $[-1,1]$.

\section{Smoke Removal in Linear RAW}
\label{sec:suppl_raw}\label{sec:suppl_operator}

In Sec.~\ref{sec:removal} of the main paper, we remove the smoke in linear RAW.
Three properties of RAW support this: the formation is linear, the output stays physical, and the noise is known.
We give the evidence for each and the parts of the operator that follow from it.

\paragraph{Linearity.}
The display encoding $\Gamma$ of Eq.~\eqref{eq:loss} compresses RAW values.
A small change $dJ$ of the clean radiance changes the hazy RAW value by $T\,dJ$, the developed hazy value by $\Gamma'(I)\,T\,dJ$, and the developed clean value by $\Gamma'(J)\,dJ$.
In RAW, the smoke therefore scales the contrast of every pixel by the same $T$.
In the developed image, it scales the contrast by the apparent transmission
\begin{equation}
    \tilde{T}(\mathbf{x}) = T\,\frac{\Gamma'\bigl(I(\mathbf{x})\bigr)}{\Gamma'\bigl(J(\mathbf{x})\bigr)},
    \label{eq:suppl_apparent_t}
\end{equation}
where we omit the channel index.
The slope $\Gamma'$ falls with brightness, so $\tilde{T}$ is lowest where the veil brightens a pixel the most (Fig.~\ref{fig:suppl_apparent_t}a).

Table~\ref{tab:suppl_domains} repeats the measurements of Sec.~\ref{sec:formation} in three domains.
In all three, Eq.~\eqref{eq:formation} fits every view better than Eq.~\eqref{eq:asm}.
In RAW, the local transmission barely changes with the brightening by the veil, whereas after developing it falls with the brightening, as Eq.~\eqref{eq:suppl_apparent_t} predicts (Fig.~\ref{fig:suppl_apparent_t}b).
Against the log brightening of each domain, its logarithm has a slope of $-0.09$ in RAW and of $-0.76$ after developing.
The detail gain in RAW therefore need not change with the brightness of the veil, and one $t$ per view sets it.
Across the rooms, the light ratio spans a factor of $5.0$ in RAW and of $1.6$ after developing.
The share to remove follows the light ratio with a rank correlation of $0.90$ in RAW and of $0.33$ after developing.
The gains $e^{\rho}$ remove this share, so in RAW their labels follow the light ratio.
Our simulator ties the level of the veil to the light of each scene (Sec.~\ref{sec:suppl_simulation}).
The network therefore predicts the gains and $t$ in linear RAW.
The same linearity separates the two layers of Eq.~\eqref{eq:operator}: the smooth veil falls into the base, and the detail is only attenuated.

\begin{figure}[t]
    \centering
    \includegraphics[width=\linewidth]{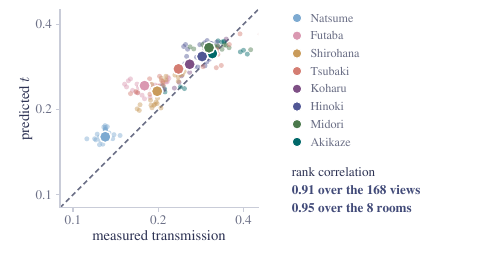}
    \caption{\textbf{Predicted and measured transmission} on the $168$ real views of Sec.~\ref{sec:formation}, as luminance averages. The predictions are those of the model of Table~\ref{tab:main_results}. The measurements come from the paired clean captures: the single $T$ per view of the fit of Eq.~\eqref{eq:formation} on the $6\times9$ grid, not the window-based $\tau$ of Table~\ref{tab:suppl_rooms}. Small dots are views, large dots room medians, and the dashed line marks equality.}
    \label{fig:suppl_t_measured}
\end{figure}

\paragraph{A physical output.}
Block means and upsampling are linear, so in RAW they act on the two terms of Eq.~\eqref{eq:formation} separately.
Let $G^I$, $G^J$, and $G^V$ be the block means of the hazy view, the clean view, and the veil, and $D^I$ and $D^J$ the details of the two views.
Then
\begin{equation}
    \begin{gathered}
        G^I_c = T_c\,G^J_c + G^V_c, \\
        D^I_c = T_c\,D^J_c + \bigl(V_c - \mathcal{U}(G^V_c)\bigr),
    \end{gathered}
    \label{eq:suppl_layers}
\end{equation}
and the last term is small where the veil is smooth within a block.
With the labels $\rho^{\star}=\log(G^J/G^I)$ and $t^{\star}=T$, and with $w_b=1$, Eq.~\eqref{eq:operator} becomes
\begin{equation}
    \hat{J}_c = \mathcal{U}(G^J_c) + T_c^{-1}\,D^I_c \approx \mathcal{U}(G^J_c) + D^J_c = J_c,
    \label{eq:suppl_inverse}
\end{equation}
the light that the camera would have recorded without smoke.
The approximation needs the last term of Eq.~\eqref{eq:suppl_layers}, multiplied by $T_c^{-1}$, to stay small against $D^J_c$.
The encoding of a developed image is nonlinear, so neither Eq.~\eqref{eq:suppl_layers} nor Eq.~\eqref{eq:suppl_inverse} holds there.
The output is therefore physical to the extent that the network predicts its labels.
Learned through the image term alone, $t$ settles near $0.5$, while its true value is $0.23$ (Sec.~\ref{sec:suppl_t_supervision}).
On the $168$ real views, Fig.~\ref{fig:suppl_t_measured} compares the predicted $t$ with the transmission that Sec.~\ref{sec:formation} measures from the paired clean capture.
Their rank correlation is $0.91$ over the views and $0.95$ over the rooms.

\begin{figure}[t]
    \centering
    \includegraphics[width=\linewidth]{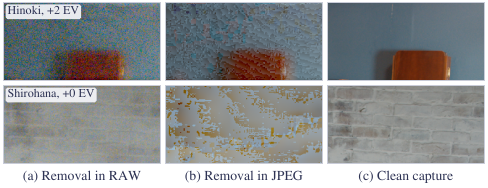}
    \caption{\textbf{The same removal in linear RAW and in the camera JPEG}, on crops of the two rooms in which the fewest $8$-bit levels remain, at about four times our working resolution. The JPEG is first mapped back to linear values. Both removals use the transmission and the veil measured from the paired clean capture, not the output of our network. The labels give the exposure added for display.}
    \label{fig:suppl_raw_vs_jpeg}
\end{figure}

\paragraph{Known noise.}
The detail gain amplifies the detail and, with it, every error of the input.
In linear RAW, this error is zero-mean sensor noise whose variance is affine in the signal~\citeS{brooks2019unprocessing,mildenhall2022rawnerf}.
The camera JPEG can be mapped back to linear values, yet it holds only $8$ bits.
Behind the smoke, the light transmitted from the scene raises a pixel above the veil by a median of $2.3$ levels in \textsc{Hinoki}, $5.3$ in \textsc{Shirohana}, and between $9.2$ and $24.7$ in the other rooms.
Fig.~\ref{fig:suppl_raw_vs_jpeg} applies the same removal to both: the amplified error is fine grain in RAW and blocks of false color in the JPEG.

The operator therefore amplifies the detail only where it rises above the noise.
The detail layer splits into three bands with Gaussian filters $\phi_\sigma$ of standard deviation $\sigma$ pixels,
\begin{equation}
    \begin{gathered}
        D_1 = \mathbf{z} - \phi_1\ast\mathbf{z},\qquad
        D_2 = \phi_1\ast\mathbf{z} - \phi_2\ast\mathbf{z},\\
        D_3 = \phi_2\ast\mathbf{z} - \mathcal{U}(G),
    \end{gathered}
    \label{eq:suppl_bands}
\end{equation}
so that $D_1 + D_2 + D_3 = D$.
We measure the noise $\sigma_c^2 = a_c\,z^{\mathrm{wb}}_c + b_c$ of the white-balanced view $z^{\mathrm{wb}}$ for every room and channel on the hazy RAW captures of that room.
The weight of band $b$ compares the local energy of the band with the energy that this noise alone would leave in it,
\begin{equation}
    w_b = \Bigl[\,1 - k_b\,\frac{\sigma^2}{E_b}\Bigr]_0^1,
    \qquad E_b = \phi_2 \ast \bigl(Y(D_b)\bigr)^2,
    \label{eq:suppl_wb}
\end{equation}
where $[\cdot]_0^1$ clips to $[0,1]$, $Y(D_b)$ is the luminance of the band, and $\sigma^2$ is the noise variance of the luminance of $\mathbf{z}$,
\begin{equation}
    \sigma^2 = s^2 \sum_c y_c^2\,\bigl(a_c\,z^{\mathrm{wb}}_c + b_c\bigr),
    \label{eq:suppl_var}
\end{equation}
where $s$ scales $z^{\mathrm{wb}}$ to $\mathbf{z}$, and $y_c$ are the Rec.~709 luminance weights of the camera.
The constants $k_b = 0.761$, $0.036$, and $0.012$ are the variances that white noise of unit variance leaves in the three bands.
For each fold, $a_c$ and $b_c$ are the medians of the per-room fits over the rooms other than the one under evaluation, scaled like the training noise.
Where a band holds only noise, $w_b\approx0$ and the band keeps its recorded amplitude; where texture dominates, $w_b$ approaches $1$ and the band receives the full gain.

\begin{figure}[t]
    \centering
    \includegraphics[width=\linewidth]{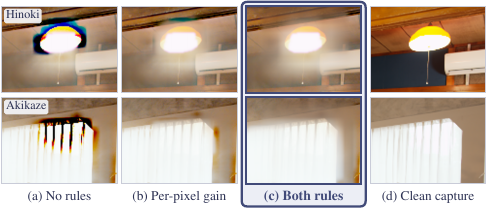}
    \caption{\textbf{The two rules next to light sources}, on crops around a lamp and a window. (a) Without the rules, a dark rim follows the lamp and streaks follow the window frame. (b) The per-pixel gain removes both. (c) Adding the detail-gain mask removes the tinted patches that remain. (d) The clean capture. All outputs use the weights $w_b$ of Eq.~\eqref{eq:suppl_wb}.}
    \label{fig:suppl_rules}
\end{figure}

\paragraph{Light sources.}
Next to light sources, the split into a smooth base and a detail layer breaks down, so we apply two rules there (Fig.~\ref{fig:suppl_rules}).
Their masks $m_r$ are built from the pixels that are clipped or brighter than four times the median: we dilate this set by $r$ pixels, blur it with a Gaussian of $\sigma=2$ pixels, and keep the original pixels at $1$.
The block mean of a source is far brighter than the pixels around the source, and bilinear upsampling spreads the change of this mean onto these pixels, which can push them below zero.
The first rule, the per-pixel gain, therefore multiplies each pixel within $m_8$ by the upsampled gain,
\begin{equation}
    \hat{J} \leftarrow \hat{J} + m_8\,\bigl(\mathbf{z}\odot\mathcal{U}(e^{\rho}) - \hat{J}\bigr),
\end{equation}
which keeps the output non-negative.
The detail gain would also amplify the edges of the source into halos.
The second rule, the detail-gain mask, therefore withholds it within $m_{24}$, $\mathbf{g} \leftarrow 1 + (\mathbf{g}-1)\,(1-m_{24})$.

\paragraph{Input and output.}
In the input, pixels with a channel at or above $0.97$ of its white level count as clipped and are set to a neutral white.
After the operator, we clamp $\hat{J}$ at zero, divide it by $s$ and by the white-balance gains, and set the clipped pixels back to the white level.

\section{Fusing the Views in 3D}
\label{sec:suppl_3d}

Gaussian splatting fuses the views that the operator corrects one by one (Sec.~\ref{sec:removal}).

\begin{figure}[t]
    \centering
    \includegraphics[width=\linewidth]{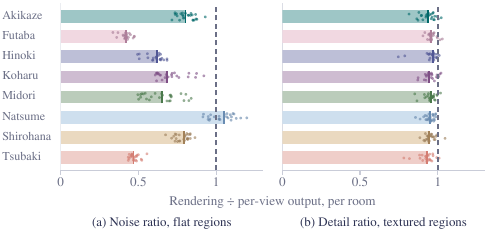}
    \caption{\textbf{Noise and detail after the reconstruction.} Ratio of the rendering to the corrected view (labeled per-view output) at every training pose: (a) noise in flat regions and (b) $2$--$4$\,px detail in textured regions. Dots are training poses, bars are room medians, and the dashed line marks a ratio of $1$.}
    \label{fig:suppl_noise_recon}
\end{figure}

\paragraph{Averaging the amplified noise.}
The detail gain amplifies the sensor noise together with the detail.
The operator only rescales the recorded values, so this noise stays zero-mean in linear RAW, and the reconstruction averages it across the views of a scene~\citeS{mildenhall2022rawnerf}.
A denoiser applied to each view before the reconstruction oversmooths the renderings~\citeS{mildenhall2022rawnerf}.
At the median over the rooms, the renderings carry $0.67$ times the noise of the corrected views and $0.95$ times their $2$--$4$\,px detail, and the noise falls in seven of the eight rooms (Fig.~\ref{fig:suppl_noise_recon}).
In every room, the renderings have higher PSNR and SSIM against the clean captures than the corrected views. Their average LPIPS is $0.254$, against $0.248$ for the corrected views (Table~\ref{tab:suppl_consistency}).

\begin{table}[!b]
    \centering
    \caption{\textbf{Corrected views and their reconstruction.} Top: against the clean captures, scored as in Table~\ref{tab:main_results}. Bottom: renderings against the corrected views, both developed as in Table~\ref{tab:main_results}. Scores of renderings are averaged over three seeds.}
    \label{tab:suppl_consistency}
    \footnotesize
    \setlength{\tabcolsep}{0pt}
    \begin{tabular*}{\linewidth}{@{}l@{\extracolsep{\fill}}ccc@{}}
        \toprule
        & PSNR$\uparrow$ & SSIM$\uparrow$ & LPIPS$\downarrow$ \\
        \midrule
        Corrected views, training poses & 18.69 & 0.734 & 0.248 \\
        Renderings, training poses & 18.89 & 0.772 & 0.254 \\
        Renderings, novel views & 18.88 & 0.759 & 0.268 \\
        \midrule
        Renderings against corrected views & 30.94 & 0.919 & -- \\
        \bottomrule
    \end{tabular*}
\end{table}

\paragraph{Removal per view.}
The physics-based methods of Table~\ref{tab:main_results} remove the smoke inside the reconstruction, where the scene depth and the agreement between the views can guide the removal.
Depth explains little of the veil, which follows the light sources (Sec.~\ref{sec:formation}).
A veil that grows with depth fails also when the removal runs per view: trained on it, our network reaches $13.87$\,dB and leaves a veil around the lamp and the windows (Table~\ref{tab:ablation}, Fig.~\ref{fig:suppl_sim_veil}).
The novel views, which the Gaussians never see, reach the PSNR of the training poses (Table~\ref{tab:suppl_consistency}).
We therefore remove the smoke view by view and use the reconstruction to fuse the corrected views.

\section{Measurement Details for \texorpdfstring{Sec.~\ref{sec:formation}}{Sec. 3.1}}
\label{sec:suppl_measurement}

The measurements of Sec.~\ref{sec:formation} use the white-balanced RAW captures of the $168$ training poses at which the lighting did not change between the hazy and the clean capture, at the working resolution of Sec.~\ref{sec:suppl_evaluation}. At the other $27$ of the $195$ training poses, the difference between the two captures does not come from the smoke alone. Scores against the clean captures at the training poses use all $195$.
Table~\ref{tab:suppl_rooms} lists the measurements per room.

\begin{table}[!b]
    \centering
    \caption{\textbf{Per-room measurements of Sec.~\ref{sec:formation}.} Advantage: median advantage of the fit of Eq.~\eqref{eq:formation} over that of Eq.~\eqref{eq:asm}, in dB. Light: share of the light in the variance of the log veil, in \%. Angle: median angle between $\mathbf{u}_1$ and the light color multiplied by $1-T_c$. $\ell$, $\tau$, $q$: room medians, as in Fig.~\ref{fig:formation}, with $\tau$ from the window regressions of this section.}
    \label{tab:suppl_rooms}
    \footnotesize
    \setlength{\tabcolsep}{0pt}
    \begin{tabular*}{\linewidth}{@{}l@{\extracolsep{\fill}}ccccccc@{}}
        \toprule
        Room & Views & Advantage & Light & Angle & $\ell$ & $\tau$ & $q$ \\
        \midrule
        \textsc{Akikaze}   & 17 & 5.3 & 80 & $0.7^\circ$ & 1.11 & 1.07 & 0.50 \\
        \textsc{Futaba}    & 23 & 6.6 & 74 & $0.3^\circ$ & 0.51 & 1.62 & 0.24 \\
        \textsc{Hinoki}    & 22 & 4.4 & 80 & $9.3^\circ$ & 2.31 & 1.45 & 0.82 \\
        \textsc{Koharu}    & 11 & 6.9 & 84 & $1.2^\circ$ & 0.52 & 1.01 & 0.20 \\
        \textsc{Midori}    & 21 & 5.1 & 92 & $2.2^\circ$ & 0.46 & 1.13 & 0.25 \\
        \textsc{Natsume}   & 25 & 6.1 & 74 & $0.5^\circ$ & 0.57 & 1.80 & 0.30 \\
        \textsc{Shirohana} & 26 & 9.3 & 92 & $1.8^\circ$ & 1.28 & 1.46 & 0.62 \\
        \textsc{Tsubaki}   & 23 & 5.3 & 70 & $1.0^\circ$ & 0.78 & 1.15 & 0.34 \\
        \bottomrule
    \end{tabular*}
\end{table}

\begin{figure}[t]
    \centering
    \includegraphics[width=\linewidth]{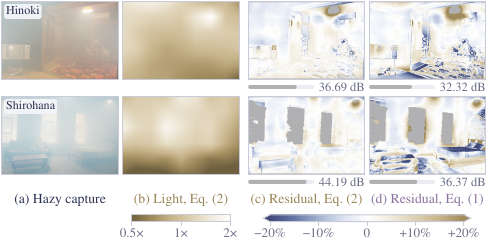}
    \caption{\textbf{Where the two fits fail.} (a) The hazy capture. (b) The light $a(\mathbf{x})$ of the fit of Eq.~\eqref{eq:formation}, relative to its median. (c), (d) The residuals of the fits of Eqs.~\eqref{eq:formation} and~\eqref{eq:asm}, positive where the capture is brighter than the fit and white where the fit is perfect. Gray marks pixels left out of the fits, and the bars give the held-out RAW PSNR, with full bars at $50$\,dB.}
    \label{fig:suppl_fit_residuals}
\end{figure}

\begin{table*}[t]
    \centering
    \caption{\textbf{Per-scene novel-view results for the operator and input ablations in Table~\ref{tab:ablation}.} We use the novel-view metrics and FujinSplat scores from Table~\ref{tab:main_results}. Detail gives the $2$--$4$\,px band amplitude in the corrected training views relative to that in the paired clean captures, with $1$ denoting equal amplitudes. Colors denote \colorbox{fst}{first}, \colorbox{sed}{second}, and \colorbox{thd}{third} place for every column and metric except Detail, here and in Tables~\ref{tab:suppl_retrained} and~\ref{tab:suppl_2d_benchmarks}. Ranks use unrounded values.}
    \label{tab:suppl_ablation_per_scene}
    \fontsize{3.0}{3.3}\selectfont
    \setlength{\tabcolsep}{3.63pt}
    \renewcommand{\arraystretch}{0.82}
    \begingroup
    \setlength{\arrayrulewidth}{0.20pt}
    \renewcommand{\colorbox}[2]{\cellcolor{#1}#2}
    \newcommand{\metricrowstrut}{\rule[-0.64ex]{0pt}{2.36ex}}
    \newcommand{\metricHeader}{\makebox[4.1em][l]{\bfseries Metrics}}
    \newcommand{\metricPSNR}{\metricrowstrut\makebox[4.1em][l]{PSNR\hfill\scalebox{0.55}{$\uparrow$}}}
    \newcommand{\metricSSIM}{\metricrowstrut\makebox[4.1em][l]{SSIM\hfill\scalebox{0.55}{$\uparrow$}}}
    \newcommand{\metricLPIPS}{\metricrowstrut\makebox[4.1em][l]{LPIPS\hfill\scalebox{0.55}{$\downarrow$}}}
    \newcommand{\metricDetail}{\metricrowstrut\makebox[4.1em][l]{Detail}}
    \resizebox{\textwidth}{!}{%
    \begin{tabular}{l|c|*{9}{r}}
        \hline
        \rule{0pt}{2.2ex}\textbf{Variant} & \metricHeader
        & \textbf{Aki.} & \textbf{Fut.} & \textbf{Hin.} & \textbf{Koh.} & \textbf{Mid.} & \textbf{Nat.} & \textbf{Shi.} & \textbf{Tsu.} & \textbf{Avg.} \\
        \hline
         & \metricPSNR & \colorbox{fst}{\textbf{20.84}} & \colorbox{fst}{\textbf{20.05}} & \colorbox{fst}{\textbf{16.95}} & \colorbox{sed}{19.13} & \colorbox{sed}{20.70} & \colorbox{sed}{17.29} & \colorbox{fst}{\textbf{17.74}} & \colorbox{thd}{18.32} & \colorbox{fst}{\textbf{18.88}} \\
        FujinSplat & \metricSSIM & \colorbox{fst}{\textbf{0.786}} & \colorbox{fst}{\textbf{0.830}} & \colorbox{fst}{\textbf{0.611}} & \colorbox{fst}{\textbf{0.779}} & \colorbox{fst}{\textbf{0.839}} & \colorbox{fst}{\textbf{0.760}} & \colorbox{fst}{\textbf{0.733}} & \colorbox{fst}{\textbf{0.732}} & \colorbox{fst}{\textbf{0.759}} \\
         & \metricLPIPS & \colorbox{fst}{\textbf{0.231}} & \colorbox{fst}{\textbf{0.225}} & \colorbox{fst}{\textbf{0.409}} & \colorbox{fst}{\textbf{0.257}} & \colorbox{fst}{\textbf{0.232}} & \colorbox{fst}{\textbf{0.250}} & \colorbox{fst}{\textbf{0.271}} & \colorbox{fst}{\textbf{0.268}} & \colorbox{fst}{\textbf{0.268}} \\
         & \metricDetail & 0.97 & 0.99 & 0.94 & 1.16 & 1.36 & 1.16 & 0.85 & 1.05 & 1.06 \\
        \cline{1-11}
         & \metricPSNR & \colorbox{thd}{19.82} & 18.39 & \colorbox{thd}{16.79} & 17.98 & 19.48 & \colorbox{thd}{15.95} & 15.90 & \colorbox{fst}{\textbf{19.24}} & 17.94 \\
        No base--detail split & \metricSSIM & 0.609 & 0.674 & 0.485 & 0.649 & 0.712 & 0.610 & 0.423 & 0.685 & 0.606 \\
         & \metricLPIPS & 0.514 & 0.498 & 0.623 & 0.459 & 0.438 & 0.521 & 0.611 & 0.490 & 0.519 \\
         & \metricDetail & 0.31 & 0.30 & 0.22 & 0.31 & 0.34 & 0.23 & 0.17 & 0.30 & 0.27 \\
        \cline{1-11}
         & \metricPSNR & \colorbox{sed}{20.38} & \colorbox{thd}{18.55} & \colorbox{sed}{16.90} & \colorbox{thd}{18.42} & \colorbox{thd}{19.70} & 15.93 & \colorbox{thd}{16.27} & \colorbox{sed}{18.73} & \colorbox{thd}{18.11} \\
        Base gain only & \metricSSIM & 0.660 & 0.697 & 0.526 & 0.693 & 0.748 & 0.625 & 0.485 & \colorbox{thd}{0.700} & 0.642 \\
         & \metricLPIPS & 0.405 & 0.435 & 0.522 & 0.390 & 0.378 & 0.488 & 0.498 & 0.423 & 0.442 \\
         & \metricDetail & 0.35 & 0.31 & 0.29 & 0.35 & 0.40 & 0.22 & 0.19 & 0.35 & 0.31 \\
        \cline{1-11}
         & \metricPSNR & 15.73 & 15.53 & 15.14 & 16.71 & 16.53 & 13.16 & 12.57 & 17.40 & 15.35 \\
        Detail gain only & \metricSSIM & \colorbox{sed}{0.692} & \colorbox{thd}{0.717} & \colorbox{thd}{0.543} & \colorbox{thd}{0.731} & \colorbox{thd}{0.754} & \colorbox{thd}{0.637} & \colorbox{sed}{0.630} & 0.686 & \colorbox{thd}{0.674} \\
         & \metricLPIPS & \colorbox{sed}{0.301} & \colorbox{thd}{0.327} & \colorbox{sed}{0.421} & \colorbox{sed}{0.286} & \colorbox{sed}{0.290} & \colorbox{sed}{0.343} & \colorbox{sed}{0.354} & \colorbox{sed}{0.298} & \colorbox{sed}{0.328} \\
         & \metricDetail & 1.69 & 1.61 & 1.82 & 1.16 & 1.31 & 2.07 & 1.84 & 1.24 & 1.59 \\
        \cline{1-11}
         & \metricPSNR & 19.01 & \colorbox{sed}{18.90} & 16.23 & \colorbox{fst}{\textbf{19.38}} & \colorbox{fst}{\textbf{20.97}} & \colorbox{fst}{\textbf{17.38}} & \colorbox{sed}{17.39} & 18.24 & \colorbox{sed}{18.44} \\
        Camera JPEG & \metricSSIM & \colorbox{thd}{0.690} & \colorbox{sed}{0.796} & \colorbox{sed}{0.580} & \colorbox{sed}{0.757} & \colorbox{sed}{0.824} & \colorbox{sed}{0.739} & \colorbox{thd}{0.620} & \colorbox{sed}{0.730} & \colorbox{sed}{0.717} \\
         & \metricLPIPS & \colorbox{thd}{0.397} & \colorbox{sed}{0.302} & \colorbox{thd}{0.452} & \colorbox{thd}{0.341} & \colorbox{thd}{0.303} & \colorbox{thd}{0.345} & \colorbox{thd}{0.442} & \colorbox{thd}{0.317} & \colorbox{thd}{0.362} \\
         & \metricDetail & 0.66 & 1.05 & 0.85 & 0.88 & 1.15 & 1.65 & 0.84 & 0.98 & 1.01 \\
        \hline
    \end{tabular}%
    }
    \endgroup
\end{table*}

\paragraph{Two models of the veil.}
The two fits test whether the smoke varies in its transmission or in its light.
We align both captures by a global sub-pixel shift and leave out clipped pixels and their $2$-pixel neighborhoods.
Both fits use $T_c = e^{-\kappa_c\tau}$ with the spectral ratio measured below and a light $A_c = u_c\,a$ of one color $u$ per view.
The fit of Eq.~\eqref{eq:formation} varies $a(\mathbf{x})$ over the view, and the fit of Eq.~\eqref{eq:asm} varies $\tau(\mathbf{x})$.
The varying quantity is interpolated in log space from the $6\times9$ grid, so each fit has $57$ parameters.
We fit to one half of the pixels, in alternating $8\times8$-pixel cells, and score by the RAW PSNR of the other half.
Eq.~\eqref{eq:formation} fits every view better, by at least $0.7$\,dB (Fig.~\ref{fig:suppl_fit_residuals}).
Freeing both maps improves the fit of Eq.~\eqref{eq:formation} by a median of $1.7$\,dB and that of Eq.~\eqref{eq:asm} by $7.9$\,dB.

\paragraph{Light and opacity in the veil.}
With both maps free, the veil $(1-T)\,A$ is the product of the light $a(\mathbf{x})$ and the opacity $1-T(\mathbf{x})$.
We split the variance of its logarithm over the grid nodes between the two, with half of their covariance each.
The remaining variation of $T$ follows the path length $r$ along the viewing ray, from monocular depth~\citeS{yang2024depthv2} aligned to the COLMAP points of the view.
The slope of $\log\tau$ against $\log r$ has a median of $0.95$, close to the $1$ of a homogeneous medium.
In the fit of Eq.~\eqref{eq:asm}, which must explain the veil by $T$ alone, it is $2.95$.

\begin{figure*}[p]
    \centering
    \includegraphics[width=\textwidth]{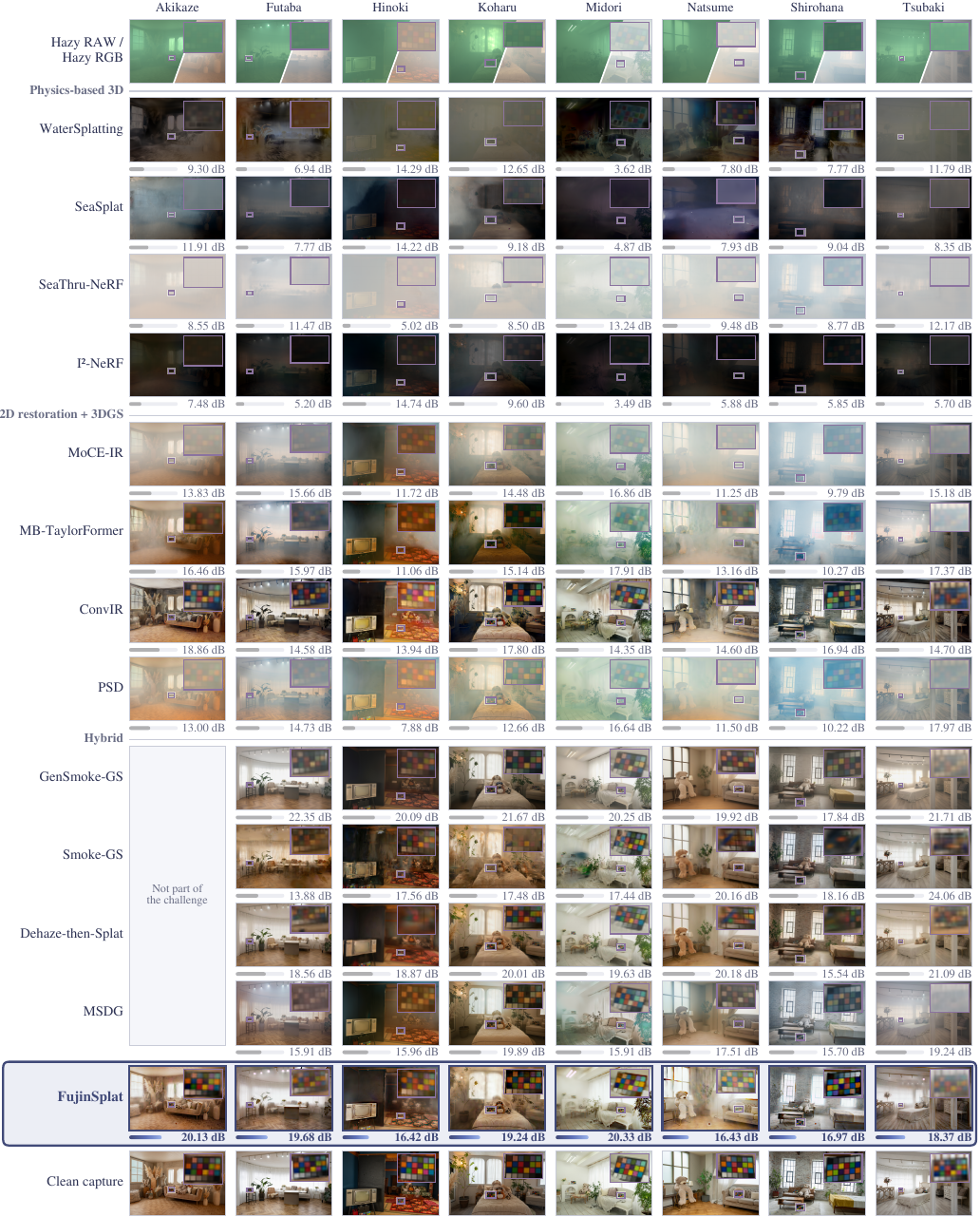}
    \caption{\textbf{Novel views of all eight RealX3D smoke scenes}, with boxes and bars as in Fig.~\ref{fig:qualitative}.}
    \label{fig:suppl_qualitative_all}
\end{figure*}

\begin{figure*}[t]
    \centering
    \includegraphics[width=\textwidth]{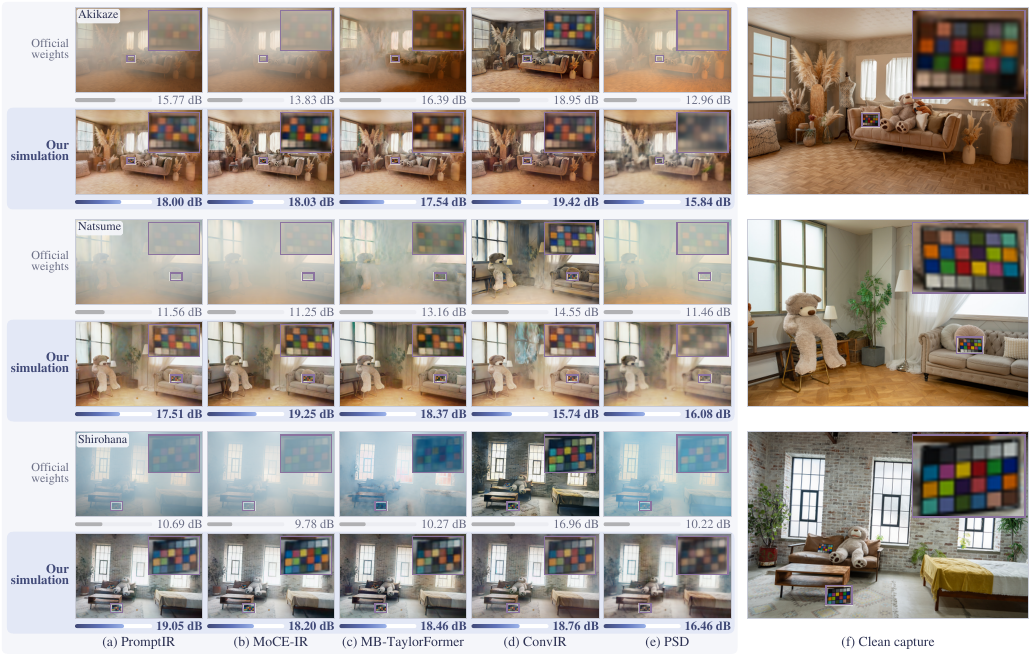}
    \caption{\textbf{Restoration networks before and after training on our simulation.} (a)--(e) The five networks with their official weights (upper rows) and after training on our simulation (lower rows). (f) The clean capture. The views, the enlarged charts, and the PSNR bars are as in Fig.~\ref{fig:qualitative}.}
    \label{fig:suppl_sim_effect}
\end{figure*}

\paragraph{One color direction.}
The singular value decomposition of Sec.~\ref{sec:formation} runs over the pixels below $0.97$ of the white level in both captures, and $\mathbf{u}_1$ is its leading direction.

\paragraph{Color of the light.}
We estimate $\kappa$ on the $202$ color charts visible in $157$ views.
An affine fit $I_c = T_c\,J_c + V_c$ per channel over the patches of a chart gives its transmissions, and $\kappa$ holds the median ratios of their optical thicknesses.
The blue-to-red ratio has a 10th to 90th percentile range of $1.41$ to $1.83$, and the ratios are nearly independent of the density of the smoke.
The light color is the mean of the unclipped window and lamp pixels of the clean capture.
Without the factor $1-T_c$, its median angle to $\mathbf{u}_1$ is $4.0^\circ$.
With this factor, the angle is largest in \textsc{Hinoki}, whose lamp cores are clipped in every view.
We attribute this to these cores, which leave only the warm rim of the shade in the measured color.

\paragraph{Light ratio and share to remove.}
We estimate the light on the smoke by regressions in windows of $32\times32$ pixels, where the fit $I_c = T_c\,J_c + V_c$ with one $\tau$ and three intercepts gives $A_c = V_c/(1-T_c)$.
Windows with little texture or next to a light source, whose glow inflates $T$, take $\tau$ from the textured ones.
The highlights excluded from $E_{95}$ are clipped or brighter than four times the median.
At the median pixel, $\operatorname{med}(Y_I) \approx T\operatorname{med}(Y_J) + (1-T)\,A_0$ by Eq.~\eqref{eq:formation}, hence
\begin{equation}
    q = 1 - \frac{1}{T + (1-T)\,\ell\,k}, \qquad k = \frac{E_{95}}{\operatorname{med}(Y_J)}.
    \label{eq:suppl_q}
\end{equation}
With the measured $T$, $\ell$, and $k$ of each view, Eq.~\eqref{eq:suppl_q} reproduces the measured $q$ with $R^2 = 0.985$.
With $T$ and $k$ fixed at their medians, $\ell$ alone explains $88\%$ of the variance of $q$, and $T$ or $k$ alone explains none of it.
The share follows $\ell$ because $\ell$ varies most: the standard deviation across views is $0.54$ for $\log\ell$, $0.10$ for $\log(1-T)$, and $0.14$ for $\log k$.
With one room left out, the Spearman correlation of $q$ with $\ell$ stays between $0.86$ and $0.96$ across rooms and between $0.88$ and $0.95$ across views, and that with $\tau$ ranges from $-0.21$ to $0.46$.

\section{Additional Results}
\label{sec:suppl_results}

Fig.~\ref{fig:qualitative} of the main paper compares FujinSplat with one family of methods per scene.
Fig.~\ref{fig:suppl_qualitative_all} shows the same views of all eight scenes with every family of methods.

\begin{table*}[t]
    \centering
    \caption{\textbf{Restoration networks trained on our simulation: novel-view results on the eight RealX3D smoke scenes}, scored as in Table~\ref{tab:main_results}. Params: number of parameters, with MB-TaylorFormer-L, ConvIR-B, and the MSBDN backbone of PSD. Time: correction time per view at the working resolution on one NVIDIA A100, in FP32 with batch size $1$ (median of $100$ runs), including the operator for FujinSplat. Official: average with the official weights, from Table~\ref{tab:main_results}. The row of FujinSplat repeats Table~\ref{tab:main_results}.}
    \label{tab:suppl_retrained}
    \fontsize{3.0}{3.3}\selectfont
    \setlength{\tabcolsep}{2.161pt}
    \renewcommand{\arraystretch}{0.82}
    \begingroup
    \setlength{\arrayrulewidth}{0.20pt}
    \renewcommand{\colorbox}[2]{\cellcolor{#1}#2}
    \newcommand{\metricrowstrut}{\rule[-0.64ex]{0pt}{2.36ex}}
    \newcommand{\metricHeader}{\makebox[4.1em][l]{\bfseries Metrics}}
    \newcommand{\metricPSNR}{\metricrowstrut\makebox[4.1em][l]{PSNR\hfill\scalebox{0.55}{$\uparrow$}}}
    \newcommand{\metricSSIM}{\metricrowstrut\makebox[4.1em][l]{SSIM\hfill\scalebox{0.55}{$\uparrow$}}}
    \newcommand{\metricLPIPS}{\metricrowstrut\makebox[4.1em][l]{LPIPS\hfill\scalebox{0.55}{$\downarrow$}}}
    \resizebox{\textwidth}{!}{%
    \begin{tabular}{l|c|c|c|*{9}{r}|r}
        \hline
        \rule{0pt}{2.2ex}\textbf{Methods} & \textbf{Params (M)} & \textbf{Time (ms)} & \metricHeader
        & \textbf{Aki.} & \textbf{Fut.} & \textbf{Hin.} & \textbf{Koh.} & \textbf{Mid.} & \textbf{Nat.} & \textbf{Shi.} & \textbf{Tsu.} & \textbf{Avg.} & \textbf{Official} \\
        \hline
         & & & \metricPSNR & 17.31 & 17.40 & 13.54 & 17.15 & \colorbox{thd}{19.78} & \colorbox{thd}{18.15} & \colorbox{sed}{18.80} & \colorbox{thd}{17.92} & \colorbox{thd}{17.51} & 14.06 \\
        PromptIR~\citeS{promptir} & 35.6 & 184 & \metricSSIM & 0.764 & 0.812 & \colorbox{thd}{0.576} & \colorbox{thd}{0.815} & \colorbox{sed}{0.842} & \colorbox{sed}{0.797} & \colorbox{fst}{\textbf{0.754}} & \colorbox{sed}{0.775} & \colorbox{sed}{0.767} & 0.587 \\
         & & & \metricLPIPS & 0.253 & 0.253 & \colorbox{fst}{\textbf{0.368}} & \colorbox{thd}{0.221} & \colorbox{sed}{0.220} & 0.229 & \colorbox{sed}{0.256} & \colorbox{sed}{0.236} & \colorbox{sed}{0.255} & 0.406 \\
        \cline{1-14}
         & & & \metricPSNR & \colorbox{thd}{18.28} & \colorbox{sed}{18.34} & 13.53 & 17.32 & \colorbox{fst}{\textbf{21.67}} & \colorbox{fst}{\textbf{19.25}} & \colorbox{fst}{\textbf{18.80}} & \colorbox{fst}{\textbf{18.63}} & \colorbox{sed}{18.23} & 14.70 \\
        MoCE-IR~\citeS{moceir} & 25.4 & 180 & \metricSSIM & \colorbox{sed}{0.796} & \colorbox{fst}{\textbf{0.833}} & 0.575 & \colorbox{fst}{\textbf{0.828}} & \colorbox{fst}{\textbf{0.859}} & \colorbox{fst}{\textbf{0.820}} & \colorbox{sed}{0.751} & \colorbox{fst}{\textbf{0.780}} & \colorbox{fst}{\textbf{0.780}} & 0.588 \\
         & & & \metricLPIPS & \colorbox{sed}{0.222} & \colorbox{sed}{0.236} & 0.418 & \colorbox{fst}{\textbf{0.206}} & \colorbox{fst}{\textbf{0.193}} & \colorbox{fst}{\textbf{0.196}} & \colorbox{fst}{\textbf{0.250}} & \colorbox{fst}{\textbf{0.232}} & \colorbox{fst}{\textbf{0.244}} & 0.417 \\
        \cline{1-14}
         & \colorbox{sed}{} & & \metricPSNR & 16.62 & 13.11 & \colorbox{sed}{14.06} & \colorbox{thd}{18.00} & 19.49 & \colorbox{sed}{18.50} & \colorbox{thd}{18.00} & 17.46 & 16.91 & 15.37 \\
        MB-TaylorFormer~\citeS{mbtaylorformer} & \colorbox{sed}{7.4} & 592 & \metricSSIM & 0.740 & 0.740 & \colorbox{sed}{0.585} & 0.808 & 0.818 & 0.795 & 0.728 & \colorbox{thd}{0.761} & 0.747 & 0.605 \\
         & \colorbox{sed}{} & & \metricLPIPS & 0.273 & 0.286 & \colorbox{sed}{0.378} & 0.262 & 0.294 & \colorbox{thd}{0.228} & 0.286 & \colorbox{thd}{0.241} & 0.281 & 0.421 \\
        \cline{1-14}
         & \colorbox{thd}{} & \colorbox{thd}{} & \metricPSNR & \colorbox{sed}{19.30} & \colorbox{thd}{17.72} & 12.95 & \colorbox{sed}{18.28} & 16.74 & 17.80 & 16.69 & 15.79 & 16.91 & 16.08 \\
        ConvIR~\citeS{convir} & \colorbox{thd}{8.6} & \colorbox{thd}{28.2} & \metricSSIM & \colorbox{fst}{\textbf{0.814}} & \colorbox{thd}{0.825} & 0.556 & \colorbox{sed}{0.827} & 0.820 & \colorbox{thd}{0.797} & 0.694 & 0.721 & 0.757 & 0.697 \\
         & \colorbox{thd}{} & \colorbox{thd}{} & \metricLPIPS & \colorbox{fst}{\textbf{0.222}} & \colorbox{thd}{0.250} & \colorbox{thd}{0.383} & \colorbox{sed}{0.215} & \colorbox{thd}{0.230} & \colorbox{sed}{0.224} & 0.299 & 0.287 & \colorbox{thd}{0.264} & 0.296 \\
        \cline{1-14}
         & & \colorbox{sed}{} & \metricPSNR & 16.05 & 14.96 & \colorbox{thd}{13.70} & 17.44 & 16.48 & 16.27 & 17.00 & 14.11 & 15.75 & 13.66 \\
        PSD~\citeS{psd} & 33.1 & \colorbox{sed}{16.2} & \metricSSIM & 0.690 & 0.752 & 0.510 & 0.739 & 0.787 & 0.718 & 0.608 & 0.698 & 0.688 & 0.605 \\
         & & \colorbox{sed}{} & \metricLPIPS & 0.418 & 0.371 & 0.502 & 0.390 & 0.352 & 0.383 & 0.444 & 0.372 & 0.404 & 0.404 \\
        \cline{1-14}
         & \colorbox{fst}{} & \colorbox{fst}{} & \metricPSNR & \colorbox{fst}{\textbf{20.84}} & \colorbox{fst}{\textbf{20.05}} & \colorbox{fst}{\textbf{16.95}} & \colorbox{fst}{\textbf{19.13}} & \colorbox{sed}{20.70} & 17.29 & 17.74 & \colorbox{sed}{18.32} & \colorbox{fst}{\textbf{18.88}} & -- \\
        \textbf{FujinSplat (Ours)} & \colorbox{fst}{\textbf{1.8}} & \colorbox{fst}{\textbf{13.6}} & \metricSSIM & \colorbox{thd}{0.786} & \colorbox{sed}{0.830} & \colorbox{fst}{\textbf{0.611}} & 0.779 & \colorbox{thd}{0.839} & 0.760 & \colorbox{thd}{0.733} & 0.732 & \colorbox{thd}{0.759} & -- \\
         & \colorbox{fst}{} & \colorbox{fst}{} & \metricLPIPS & \colorbox{thd}{0.231} & \colorbox{fst}{\textbf{0.225}} & 0.409 & 0.257 & 0.232 & 0.250 & \colorbox{thd}{0.271} & 0.268 & 0.268 & -- \\
        \hline
    \end{tabular}%
    }
    \endgroup
\end{table*}

\begin{table*}[t]
    \centering
    \caption{\textbf{Results on real paired benchmarks.} Scores of the hazy input, the five restoration networks of Table~\ref{tab:main_results}, and the two FujinSplat variants.}
    \label{tab:suppl_2d_benchmarks}
    \fontsize{3.0}{3.3}\selectfont
    \setlength{\tabcolsep}{2.52pt}
    \renewcommand{\arraystretch}{0.82}
    \begingroup
    \setlength{\arrayrulewidth}{0.20pt}
    \renewcommand{\colorbox}[2]{\cellcolor{#1}#2}
    \newcommand{\metricrowstrut}{\rule[-0.64ex]{0pt}{2.36ex}}
    \newcommand{\metricHeader}{\makebox[4.1em][l]{\bfseries Metrics}}
    \newcommand{\metricPSNR}{\metricrowstrut\makebox[4.1em][l]{PSNR\hfill\scalebox{0.55}{$\uparrow$}}}
    \newcommand{\metricSSIM}{\metricrowstrut\makebox[4.1em][l]{SSIM\hfill\scalebox{0.55}{$\uparrow$}}}
    \newcommand{\metricLPIPS}{\metricrowstrut\makebox[4.1em][l]{LPIPS\hfill\scalebox{0.55}{$\downarrow$}}}
    \resizebox{\textwidth}{!}{%
    \begin{tabular}{l|c|*{7}{c}}
        \hline
        \rule{0pt}{2.2ex}\textbf{Methods} & \metricHeader
        & \textbf{I-HAZE} & \textbf{O-HAZE} & \textbf{Dense-Haze} & \textbf{NH-HAZE} & \textbf{REVIDE} & \textbf{LMHaze} & \textbf{SmokeBench} \\
        \hline
         & \metricPSNR & 12.68 & 15.00 & 8.50 & 11.34 & 15.08 & 14.11 & 18.60 \\
        Hazy input & \metricSSIM & 0.676 & 0.635 & 0.424 & 0.406 & 0.755 & 0.609 & 0.763 \\
         & \metricLPIPS & 0.363 & 0.340 & 0.655 & 0.533 & 0.297 & 0.490 & 0.285 \\
        \cline{1-9}
         & \metricPSNR & 19.19 & \colorbox{thd}{24.39} & 13.87 & 17.80 & 20.57 & 16.72 & 22.21 \\
        PromptIR~\citeS{promptir} & \metricSSIM & 0.811 & \colorbox{thd}{0.845} & 0.500 & 0.623 & 0.858 & 0.701 & 0.793 \\
         & \metricLPIPS & 0.355 & \colorbox{thd}{0.257} & 0.622 & 0.405 & 0.319 & \colorbox{thd}{0.387} & 0.276 \\
        \cline{1-9}
         & \metricPSNR & 19.85 & \colorbox{fst}{\textbf{26.65}} & 14.82 & \colorbox{fst}{\textbf{18.40}} & \colorbox{thd}{21.67} & \colorbox{sed}{17.78} & \colorbox{sed}{26.19} \\
        MoCE-IR~\citeS{moceir} & \metricSSIM & 0.837 & \colorbox{fst}{\textbf{0.878}} & 0.511 & \colorbox{sed}{0.649} & \colorbox{sed}{0.882} & \colorbox{sed}{0.732} & \colorbox{sed}{0.835} \\
         & \metricLPIPS & \colorbox{fst}{\textbf{0.326}} & \colorbox{fst}{\textbf{0.229}} & 0.638 & \colorbox{sed}{0.388} & \colorbox{sed}{0.274} & \colorbox{fst}{\textbf{0.357}} & \colorbox{sed}{0.234} \\
        \cline{1-9}
         & \metricPSNR & \colorbox{fst}{\textbf{21.94}} & 22.99 & 14.84 & 18.00 & 20.53 & 16.71 & 22.68 \\
        MB-TaylorFormer~\citeS{mbtaylorformer} & \metricSSIM & \colorbox{sed}{0.859} & 0.833 & 0.485 & \colorbox{thd}{0.641} & 0.853 & \colorbox{thd}{0.704} & 0.789 \\
         & \metricLPIPS & \colorbox{thd}{0.347} & 0.268 & \colorbox{thd}{0.618} & \colorbox{thd}{0.402} & 0.308 & 0.393 & 0.285 \\
        \cline{1-9}
         & \metricPSNR & \colorbox{sed}{21.06} & 23.23 & \colorbox{sed}{16.16} & 17.98 & \colorbox{sed}{22.01} & \colorbox{thd}{17.41} & \colorbox{fst}{\textbf{26.57}} \\
        ConvIR~\citeS{convir} & \metricSSIM & \colorbox{fst}{\textbf{0.865}} & \colorbox{sed}{0.855} & \colorbox{fst}{\textbf{0.576}} & \colorbox{fst}{\textbf{0.654}} & \colorbox{fst}{\textbf{0.885}} & \colorbox{fst}{\textbf{0.733}} & \colorbox{fst}{\textbf{0.841}} \\
         & \metricLPIPS & 0.358 & \colorbox{fst}{\textbf{0.229}} & \colorbox{fst}{\textbf{0.586}} & \colorbox{fst}{\textbf{0.387}} & \colorbox{fst}{\textbf{0.263}} & \colorbox{sed}{0.370} & \colorbox{fst}{\textbf{0.231}} \\
        \cline{1-9}
         & \metricPSNR & \colorbox{thd}{20.91} & 22.71 & 15.26 & 18.18 & 20.55 & 16.32 & 22.17 \\
        PSD~\citeS{psd} & \metricSSIM & 0.810 & 0.808 & 0.499 & 0.634 & 0.841 & 0.681 & 0.766 \\
         & \metricLPIPS & 0.394 & 0.319 & \colorbox{sed}{0.602} & 0.409 & 0.360 & 0.462 & 0.328 \\
        \cline{1-9}
         & \metricPSNR & 20.42 & 24.09 & \colorbox{thd}{16.11} & \colorbox{sed}{18.37} & 21.48 & 17.24 & 20.09 \\
        \smash{\begin{tabular}[c]{@{}l@{}}\textbf{FujinSplat}\\[0.5pt]$t$ from the simulation\end{tabular}} & \metricSSIM & \colorbox{thd}{0.846} & 0.825 & \colorbox{thd}{0.543} & 0.596 & 0.839 & 0.689 & 0.698 \\
         & \metricLPIPS & 0.351 & 0.289 & 0.620 & 0.458 & 0.334 & 0.417 & 0.331 \\
        \cline{1-9}
         & \metricPSNR & 20.39 & \colorbox{sed}{25.05} & \colorbox{fst}{\textbf{16.26}} & \colorbox{thd}{18.33} & \colorbox{fst}{\textbf{23.57}} & \colorbox{fst}{\textbf{18.32}} & \colorbox{thd}{23.80} \\
        \smash{\begin{tabular}[c]{@{}l@{}}\textbf{FujinSplat}\\[0.5pt]$t$ from the image loss\end{tabular}} & \metricSSIM & 0.823 & 0.822 & \colorbox{sed}{0.546} & 0.589 & \colorbox{thd}{0.859} & 0.693 & \colorbox{thd}{0.795} \\
         & \metricLPIPS & \colorbox{sed}{0.336} & 0.284 & 0.625 & 0.458 & \colorbox{thd}{0.293} & 0.419 & \colorbox{thd}{0.275} \\
        \hline
    \end{tabular}%
    }
    \endgroup
\end{table*}

\section{Operator and Input Ablations}
\label{sec:suppl_sim_input}

Table~\ref{tab:suppl_ablation_per_scene} expands the operator and input rows of Table~\ref{tab:ablation} per scene.
SSIM and LPIPS of every variant are worse than those of FujinSplat in every scene.
In \textsc{Tsubaki}, PSNR is higher without the base--detail split and with the base gain only than with the full operator.
Without the split or with the base gain only, the corrected views keep about a third of the clean detail, and with the detail gain only they exceed it.
With the camera JPEG as input, PSNR is higher than with RAW input in three scenes and lower in the other five. Behind the smoke, the scene spans a median of only $2.3$ to $24.7$ levels of the $8$-bit JPEG, and the removal amplifies these levels into blocks of false color (Sec.~\ref{sec:suppl_raw}, Fig.~\ref{fig:suppl_raw_vs_jpeg}).

\section{Networks Trained on Our Simulation}
\label{sec:suppl_retrained}

\paragraph{Setting.}
We fine-tune the five restoration networks of Table~\ref{tab:main_results} from their official weights on our simulated pairs only, cross-validated by room as in the main paper.
They receive the normalized linear RAW view $\mathbf{z}$ without the power curve of Sec.~\ref{sec:suppl_training} and output a linear RAW image.
Each keeps its official loss, optimizer, and learning rate, and the loss is computed in the display encoding $\Gamma$ of Eq.~\eqref{eq:loss}.
We use FujinSplat's training schedule, iteration count, crop size, and batch size.
For real scenes, we process the network inputs and outputs as in Sec.~\ref{sec:suppl_operator}. We then train 3DGS on the corrected views for $18{,}000$ iterations per room and develop the rendered novel views with the scene-specific ISP.

\paragraph{Effect of the simulation.}
Table~\ref{tab:suppl_retrained} and Fig.~\ref{fig:suppl_sim_effect} compare the five networks before and after training on our simulation.
All five networks improve in average PSNR and SSIM. The PSNR gains range from $0.83$\,dB for ConvIR to $3.53$\,dB for MoCE-IR. Average LPIPS decreases for four networks and is $0.404$ before and after training for PSD.
Over the five networks and eight scenes, PSNR improves in $31$ of $40$ cases, SSIM in $37$, and LPIPS in $35$.
Training on our simulation largely removes the tinted veil that remains over \textsc{Akikaze} and \textsc{Shirohana} with the official weights (Fig.~\ref{fig:suppl_sim_effect}).
The gain is smallest in \textsc{Hinoki}, the scene that requires the most removal, where PromptIR and ConvIR lose PSNR.

\paragraph{Comparison with FujinSplat.}
FujinSplat has the highest average PSNR and the highest PSNR in four scenes, and its lead over the best retrained network is largest in \textsc{Hinoki} (Table~\ref{tab:suppl_retrained}).
MoCE-IR and PromptIR have higher average SSIM than FujinSplat, and MoCE-IR, PromptIR, and ConvIR have lower average LPIPS.
MoCE-IR is better than FujinSplat in SSIM in seven scenes and in LPIPS in six.

\begin{figure}[t]
    \centering
    \includegraphics[width=\linewidth]{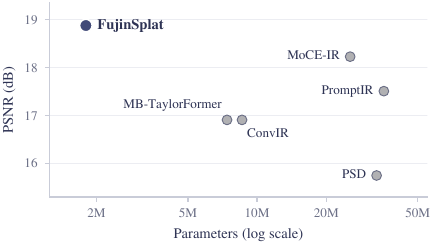}
    \caption{\textbf{Network size and novel-view PSNR} of the six networks in Table~\ref{tab:suppl_retrained}, all trained on our simulated pairs. Average PSNR over the eight scenes, against the number of parameters on a log scale.}
    \label{fig:suppl_params}
\end{figure}

\paragraph{Network size.}
In linear RAW, the effect of the smoke on a view is low-dimensional.
The residual between the hazy and the clean captures holds on average $98.7\%$ of its energy along one color direction, Eq.~\eqref{eq:formation} with one transmission per view fits every view better than Eq.~\eqref{eq:asm} (Sec.~\ref{sec:formation}), and the smooth veil falls into the base of the operator (Sec.~\ref{sec:suppl_raw}).
The predictor of FujinSplat therefore outputs three log-gains per $8\times8$ block and one transmission per view, and it has $1.8$M parameters.
The five networks of Table~\ref{tab:suppl_retrained} differ in architecture as well as in size, from $7.4$M to $35.6$M parameters, and after fine-tuning with RAW input on our simulated pairs, their ranks do not follow their sizes. Fig.~\ref{fig:suppl_params} plots the average PSNR of these five networks and FujinSplat against the number of parameters: PSD, the second largest, has the lowest PSNR, and FujinSplat, the smallest, has the highest.
From $375$ to $3{,}000$ pairs, the PSNR of FujinSplat changes by at most $0.22$\,dB (Table~\ref{tab:sensitivity}).
Our predictor follows the encoder of ConvIR, and ConvIR-B has $8.6$M parameters.
Trained on the same pairs, FujinSplat and ConvIR-B reach an SSIM of $0.759$ and $0.757$, an LPIPS of $0.268$ and $0.264$, and a PSNR of $18.88$ and $16.91$\,dB.
On one A100 at the working resolution, FujinSplat corrects a view in $13.6$\,ms, including $4.1$\,ms for the predictor, and the five networks take $16$ to $592$\,ms (Table~\ref{tab:suppl_retrained}).

\paragraph{Output form.}
FujinSplat predicts gains and one transmission, so that its output stays a linear RAW image and both quantities can be supervised with simulated labels (Sec.~\ref{sec:removal}).
It is on par with the five networks that regress pixels. Trained on the same pairs, it reaches the highest PSNR of the six, and its SSIM and LPIPS lie within the range of the five (Table~\ref{tab:suppl_retrained}).
Within this form, one gain map without the split into base and detail lowers SSIM from $0.759$ to $0.606$ and raises LPIPS from $0.268$ to $0.519$ (Table~\ref{tab:ablation}).

\section{Real Paired 2D Benchmarks}
\label{sec:suppl_2d_benchmarks}

Table~\ref{tab:suppl_2d_benchmarks} reports results on seven public benchmarks of real paired hazy and clean images: I-HAZE~\citeS{ancuti2018ihaze}, O-HAZE~\citeS{ancuti2018ohaze}, Dense-Haze~\citeS{ancuti2019dense}, NH-HAZE~\citeS{ancuti2020nhhaze}, REVIDE~\citeS{zhang2021revide}, LMHaze~\citeS{zhang2024lmhaze}, and SmokeBench~\citeS{jin2025smokebench}.
They consist of paired single images or video frames, so they test the restoration of each image without 3D reconstruction.

\begin{table}[t]
    \centering
    \caption{\textbf{Splits of the real paired benchmarks}, in image pairs, and the number of training iterations.}
    \label{tab:suppl_2d_splits}
    \footnotesize
    \setlength{\tabcolsep}{3pt}
    \begin{tabular*}{\linewidth}{@{}l@{\extracolsep{\fill}}rrrr@{}}
        \toprule
        Benchmark & Train & Val. & Test & Iterations \\
        \midrule
        I-HAZE & $22$ & $3$ & $5$ & $5{,}000$ \\
        O-HAZE & $35$ & $5$ & $5$ & $5{,}000$ \\
        Dense-Haze & $45$ & $5$ & $5$ & $5{,}000$ \\
        NH-HAZE & $45$ & $5$ & $5$ & $5{,}000$ \\
        REVIDE & $1{,}633$ & $64$ & $284$ & $10{,}000$ \\
        LMHaze & $3{,}772$ & $153$ & $1{,}115$ & $10{,}000$ \\
        SmokeBench & $9{,}775$ & $100$ & $100$ & $10{,}000$ \\
        \bottomrule
    \end{tabular*}
\end{table}

\paragraph{Protocol.}
We evaluate every method on the test sets of Table~\ref{tab:suppl_2d_splits}.
I-HAZE, O-HAZE, Dense-Haze, and NH-HAZE are split by image number, and REVIDE, LMHaze, and SmokeBench keep their official training and test sets.
The five networks start from the official weights of Table~\ref{tab:main_results} and keep their official loss, optimizer, and learning rate.
All methods use the number of iterations in Table~\ref{tab:suppl_2d_splits} and share the crop size, effective batch size, and schedule in Sec.~\ref{sec:suppl_training}, with random horizontal flips.
For each run, we select the checkpoint that performs best on the validation set and report its test-set metrics.
These benchmarks provide no label of $t$, and we report two versions of FujinSplat. Both start from the network trained on our simulation, whose $t$ was learned from the labels of the simulated pairs. The first keeps the $t$ head of this network frozen, and the second adapts it to each benchmark through the image loss. Neither version receives a label of $t$ from the benchmarks, so their comparison measures the effect of adapting $t$ to new data. The effect of the label itself is measured on the simulated pairs, where labels exist, by learning $t$ with and without it (Sec.~\ref{sec:suppl_t_supervision}).

\paragraph{Results.}
With $t$ adapted through the image loss, FujinSplat has the highest PSNR on three of the seven benchmarks, Dense-Haze, REVIDE, and LMHaze (Table~\ref{tab:suppl_2d_benchmarks}). Adapting $t$ raises its PSNR on five of the seven.
On all seven benchmarks, the highest SSIM and the lowest LPIPS come from the restoration baselines.

\section{Limitations}
\label{sec:suppl_limitations}

The main paper lists the limitations of FujinSplat. Here we show where they appear in our results.

\paragraph{Light ratio.}
The light ratio enters training through the veil level $A_0 = w\,\bar{L}$ of the simulation (Sec.~\ref{sec:simulation}), and the network does not receive it as input.
Trained on the fold of \textsc{Hinoki} and applied to all eight rooms, our network darkens them by $0.57$ to $1.27$\,EV, while they require $0.27$ to $2.06$\,EV (Fig.~\ref{fig:suppl_darkening}).
It removes too little in \textsc{Hinoki}, which requires the most.

\paragraph{Clipped lamps.}
A lamp that clips in every view hides its brightness and its color.
In \textsc{Hinoki}, whose lamp cores are clipped in every view, the measured light color keeps only the warm rim of the shade (Sec.~\ref{sec:suppl_measurement}), and a glow remains around the lamp after removal (Fig.~\ref{fig:suppl_rules}).

\paragraph{Signal below the noise.}
Where the transmitted signal falls below the noise, the noise weight withholds the detail gain, and the detail stays attenuated.
In \textsc{Shirohana}, the corrected views keep $0.85$ of the clean $2$--$4$\,px detail (Table~\ref{tab:suppl_ablation_per_scene}).

\paragraph{One camera.}
The measurements, the simulator statistics, and the noise model all come from one camera and eight rooms, since paired hazy and clean RAW captures of indoor smoke remain scarce.
Another camera needs its own noise and spectral calibration.

\section*{References}
{\small
\providecommand{\natexlab}[1]{#1}
\providecommand{\url}[1]{\texttt{#1}}
\expandafter\ifx\csname urlstyle\endcsname\relax
  \providecommand{\doi}[1]{doi: #1}\else
  \providecommand{\doi}{doi: \begingroup \urlstyle{rm}\Url}\fi
\begin{list}{[\arabic{suppref}]}{\usecounter{suppref}\settowidth{\labelwidth}{[19]}\setlength{\leftmargin}{\labelwidth}\addtolength{\leftmargin}{\labelsep}\setlength{\itemsep}{-1pt}\setlength{\parsep}{0pt}\sloppy}
\item\label{sref:ancuti2018ihaze}
Cosmin Ancuti, Codruta~O. Ancuti, Radu Timofte, and Christophe De~Vleeschouwer.
\newblock {I-HAZE}: A dehazing benchmark with real hazy and haze-free indoor
  images.
\newblock In \emph{Advanced Concepts for Intelligent Vision Systems (ACIVS)},
  pages 620--631, 2018{\natexlab{a}}.

\item\label{sref:ancuti2018ohaze}
Codruta~O. Ancuti, Cosmin Ancuti, Radu Timofte, and Christophe De~Vleeschouwer.
\newblock {O-HAZE}: A dehazing benchmark with real hazy and haze-free outdoor
  images.
\newblock In \emph{CVPRW}, 2018{\natexlab{b}}.

\item\label{sref:ancuti2019dense}
Codruta~O. Ancuti, Cosmin Ancuti, Mateu Sbert, and Radu Timofte.
\newblock {Dense-Haze}: A benchmark for image dehazing with dense-haze and
  haze-free images.
\newblock In \emph{ICIP}, pages 1014--1018, 2019.

\item\label{sref:ancuti2020nhhaze}
Codruta~O. Ancuti, Cosmin Ancuti, and Radu Timofte.
\newblock {NH-HAZE}: An image dehazing benchmark with non-homogeneous hazy and
  haze-free images.
\newblock In \emph{CVPRW}, 2020.

\item\label{sref:brooks2019unprocessing}
Tim Brooks, Ben Mildenhall, Tianfan Xue, Jiawen Chen, Dillon Sharlet, and
  Jonathan~T. Barron.
\newblock Unprocessing images for learned raw denoising.
\newblock In \emph{CVPR}, 2019.

\item\label{sref:psd}
Zeyuan Chen, Yangchao Wang, Yang Yang, and Dong Liu.
\newblock {PSD}: Principled synthetic-to-real dehazing guided by physical
  priors.
\newblock In \emph{CVPR}, pages 7176--7185, 2021.

\item\label{sref:convir}
Yuning Cui, Wenqi Ren, Xiaochun Cao, and Alois Knoll.
\newblock Revitalizing convolutional network for image restoration.
\newblock \emph{IEEE Transactions on Pattern Analysis and Machine
  Intelligence}, 46\penalty0 (12):\penalty0 9423--9438, 2024.

\item\label{sref:jin2025smokebench}
Wenzhuo Jin, Qianfeng Yang, Xianhao Wu, Hongming Chen, Pengpeng Li, and Xiang
  Chen.
\newblock {SmokeBench}: A real-world dataset for surveillance image desmoking
  in early-stage fire scenes.
\newblock In \emph{ACM MM}, pages 12722--12728, 2025.

\item\label{sref:kerbl3Dgaussians}
Bernhard Kerbl, Georgios Kopanas, Thomas Leimkuehler, and George Drettakis.
\newblock 3{D} {G}aussian splatting for real-time radiance field rendering.
\newblock \emph{ACM TOG}, 42\penalty0 (4), 2023.

\item\label{sref:kingma2015adam}
Diederik~P. Kingma and Jimmy Ba.
\newblock {Adam}: A method for stochastic optimization.
\newblock In \emph{ICLR}, 2015.

\item\label{sref:li2019benchmarking}
Boyi Li, Wenqi Ren, Dengpan Fu, Dacheng Tao, Dan Feng, Wenjun Zeng, and
  Zhangyang Wang.
\newblock Benchmarking single-image dehazing and beyond.
\newblock \emph{IEEE TIP}, 28\penalty0 (1):\penalty0 492--505, 2019.

\item\label{sref:mildenhall2022rawnerf}
Ben Mildenhall, Peter Hedman, Ricardo Martin-Brualla, Pratul~P. Srinivasan, and
  Jonathan~T. Barron.
\newblock {NeRF} in the dark: High dynamic range view synthesis from noisy raw
  images.
\newblock In \emph{CVPR}, pages 16169--16178, 2022.

\item\label{sref:promptir}
Vaishnav Potlapalli, Syed~Waqas Zamir, Salman Khan, and Fahad Shahbaz~Khan.
\newblock {PromptIR}: Prompting for all-in-one image restoration.
\newblock In \emph{NeurIPS}, pages 71275--71293, 2023.

\item\label{sref:mbtaylorformer}
Yuwei Qiu, Kaihao Zhang, Chenxi Wang, Wenhan Luo, Hongdong Li, and Zhi Jin.
\newblock {MB-TaylorFormer}: Multi-branch efficient transformer expanded by
  taylor formula for image dehazing.
\newblock In \emph{ICCV}, pages 12802--12813, 2023.

\item\label{sref:yang2024depthv2}
Lihe Yang, Bingyi Kang, Zilong Huang, Zhen Zhao, Xiaogang Xu, Jiashi Feng, and
  Hengshuang Zhao.
\newblock Depth anything {V2}.
\newblock In \emph{NeurIPS}, pages 21875--21911, 2024.

\item\label{sref:moceir}
Eduard Zamfir, Zongwei Wu, Nancy Mehta, Yuedong Tan, Danda~Pani Paudel, Yulun
  Zhang, and Radu Timofte.
\newblock Complexity experts are task-discriminative learners for any image
  restoration.
\newblock In \emph{CVPR}, pages 12753--12763, 2025.

\item\label{sref:zhang2018unreasonable}
Richard Zhang, Phillip Isola, Alexei~A. Efros, Eli Shechtman, and Oliver Wang.
\newblock The unreasonable effectiveness of deep features as a perceptual
  metric.
\newblock In \emph{CVPR}, 2018.

\item\label{sref:zhang2024lmhaze}
Ruikun Zhang, Hao Yang, Yan Yang, Ying Fu, and Liyuan Pan.
\newblock {LMHaze}: Intensity-aware image dehazing with a large-scale
  multi-intensity real haze dataset.
\newblock In \emph{ACM Int. Conf. Multimedia in Asia}, 2024.

\item\label{sref:zhang2021revide}
Xinyi Zhang, Hang Dong, Jinshan Pan, Chao Zhu, Ying Tai, Chengjie Wang, Jilin
  Li, Feiyue Huang, and Fei Wang.
\newblock Learning to restore hazy video: A new real-world dataset and a new
  method.
\newblock In \emph{CVPR}, pages 9239--9248, 2021.
\end{list}
}